%% file: ceo_icml.tex
\documentclass{article}

    \PassOptionsToPackage{square, numbers}{natbib}

\usepackage[preprint]{neurips_2022}
\usepackage{natbib}

\usepackage[utf8]{inputenc} 
\usepackage[T1]{fontenc}    
\usepackage{hyperref}       
\usepackage{url}            
\usepackage{booktabs}       
\usepackage{amsfonts}       
\usepackage{nicefrac}       
\usepackage{microtype}      
\usepackage{xcolor}         

\usepackage{algorithm}
\usepackage{algpseudocode}
\usepackage{algcompatible}

\usepackage{microtype}
\usepackage{graphicx}
\usepackage{placeins}
\usepackage{tcolorbox}
\usepackage{subcaption}
\usepackage{tabularx}
\usepackage{multirow}
\usepackage{setspace}
\usepackage{hyperref}

\usepackage{amsmath}
\usepackage{amssymb}
\usepackage{mathtools}
\usepackage{amsthm}
\usepackage{xspace}

\usepackage[capitalize]{cleveref}
\crefformat{section}{\S#2#1#3}
\crefformat{subsection}{\S#2#1#3}
\crefformat{subsubsection}{\S#2#1#3}

\theoremstyle{plain}

\theoremstyle{definition}

\theoremstyle{remark}

\usepackage[textsize=tiny]{todonotes}

\usepackage[utf8]{inputenc} 
\usepackage[T1]{fontenc}    
\hypersetup{colorlinks,linkcolor={blue},citecolor={blue},urlcolor={blue}}

\usepackage{url}            
\usepackage{nicefrac}       
\usepackage{microtype}      
\usepackage[british]{babel}
\usepackage{siunitx} 
\usepackage{tikz} 
\usepackage{tikzscale}
\usetikzlibrary{bayesnet}
\usepackage{pbox}
\usepackage{fancybox}
\usepackage{acro}
\usepackage{enumitem}
\setlist[itemize]{leftmargin=*}
\usetikzlibrary{calc}
\usepackage{float}
\usepackage{wrapfig}

\input{macros}

\input{notation}

\title{Causal Entropy Optimization}

\author{%
  Nicola Branchini  \\
  University of Edinburgh\\
  The Alan Turing Institute \\
  \And
  Virginia Aglietti \\
  University of Warwick \\
  The Alan Turing Institute \\
  \AND
  Neil Dhir \\
  The Alan Turing Institute \\
  \And
  Theodoros Damoulas \\
  University of Warwick \\
  The Alan Turing Institute \\
}

\begin{document}
\setlength{\abovedisplayskip}{0.3cm}
\setlength{\belowdisplayskip}{0.3cm}

\maketitle

\begin{abstract}
We study the problem of globally optimizing the causal effect on a target variable of an \emph{unknown causal graph} in which interventions can be performed. This problem arises in many areas of science including biology, operations research and healthcare. We propose Causal Entropy Optimization (\our), a framework which generalizes Causal Bayesian Optimization (\cbo) \cite{cbo} to account for all sources of uncertainty, including the one arising from the causal graph structure. \ceo incorporates the causal structure uncertainty \emph{both} in the surrogate models for the causal effects and in the mechanism used to select interventions via an information-theoretic acquisition function. The resulting algorithm automatically trades-off structure learning and causal effect optimization, while naturally accounting for observation noise. For various synthetic and real-world structural causal models, \ceo achieves faster convergence to the global optimum compared with \cbo while also learning the graph. Furthermore, our joint approach to structure learning and causal optimization improves upon sequential, structure-learning-first approaches.
\end{abstract}

\section{Introduction}

Causal Bayesian Networks (\cbn{s}) \citep{pearl2009causality}  offer a powerful tool for formulating and testing causal relationships among a set of random variables. Representing a system with a \cbn allows one to \eg estimate causal effects or find interventions optimizing a target node. These tasks often assume, either implicitly or explicitly, \emph{exact} knowledge of the underlying causal graph. Therefore, structure learning (also called ``causal discovery'') from data \citep{glymour2019review} has received increasing attention in the last few years. In particular, several studies have taken a Bayesian approach by formulating a prior over graphs and selecting interventions to learn the structure via Bayesian Optimal Experimental Design \citep[\boed,][]{murphy2001active, tong2001active, masegosa2013interactive, hauser2014two, kocaoglu2017cost, ness2017bayesian, von2019optimal, gamella2020active,vowels2021d}. Among these studies, a Bayesian Optimization (\bo)-based algorithm, targeted at structure learning, has been proposed by \citet{von2019optimal} with the goal of reducing the cost of discovering the true graph. A causal \bo-based algorithm \citep[\cbo,][]{cbo} was also recently developed to identify interventions maximizing a target variable, given a known causal graph -- a problem named \emph{causal global optimization}, which we refer to simply as causal optimization. 
While the two works develop \bo algorithms tackling the tasks of structure learning and causal optimization separately, the investigator is often interested in learning optimal actions while not having exact knowledge of the causal relationships among variables.
This is the challenging setting we consider in this paper which is the first focusing on solving these two problems jointly.  

\textbf{Example} Consider a setting where the investigator aims at finding the level of statin drug or aspirin that should be prescribed to a patient in order to minimize the level of prostate specific antigen (\psa).

\begin{wrapfigure}[13]{r}{0.4\textwidth}
\centering
\includegraphics[width=0.30\textwidth]{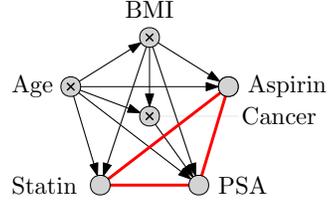}
\caption{\begin{small} Causal \DAG with unoriented edges in red. Shaded and crossed nodes represent manipulative and non-manipulative variables respectively. $\psa$ is the outcome of interest. \end{small}}
\label{fig:graph_health}
\end{wrapfigure}

While the investigator might have a good understanding of the variables affecting the level of \psa (see \cref{fig:graph_health}), she might not know the exact causal relationships among them. Therefore, multiple causal graphs might be  consistent with her domain knowledge. 
For instance, the causal relationships between age, body mass index (\bmi) and cancer might be known, but those among cancer, $\psa$ and different levels of medication administration might be unknown. This is represented by the red edges in \cref{fig:graph_health}. Identifying the optimal drug dosages with \cbo would require knowledge of these edges. Indeed, \cbo \emph{assumes} \citep[appendix, Fig. 3]{cbo} them to be oriented as $\{\text{Statin}\rightarrow \text{Cancer}, \text{Statin}\rightarrow \psa, \text{Aspirin}\rightarrow \text{Cancer}, \text{Aspirin}\rightarrow \psa  \}$. 
We propose Causal Entropy Optimization (\our), a framework which, instead of assuming a specific causal graph for the data generating process, accounts for the structure uncertainty by employing a Bayesian prior. \ceo considers the causal graph prior both in the surrogate models and the acquisition function used within the \bo algorithm. 

\textbf{Contributions} We make the following contributions: 

$\bullet$ We generalize the causal global optimization problem to settings where the graph structure is fully or partially unknown.

$\bullet$ We offer the first solution to this problem which we call Causal Entropy Optimization (\ceo). \ceo models the causal effects via a set of surrogate models that account for both graph and observation uncertainty. 

$\bullet$ We introduce an information-theoretic acquisition function, which we call \emph{causal entropy search} (\ces) that addresses the trade-off between learning the causal graph and identifying the best intervention. \ces encompasses existing acquisitions used for \bo and experimental design for structure learning as special cases, and could be adapted to solve other active learning tasks.

$\bullet$ We demonstrate across synthetic and real-world causal graphs how accounting for uncertainty leads to faster convergence to the global optimum compared to \cbo. In addition, we show how exact knowledge of the causal structure is not always needed for efficient causal global optimization.

\section{Background and Problem Statement}
\label{sec:back}

We consider a probabilistic causal model \citep{pearl2009causality} consisting of a directed acyclic graph $\mathcal{G}$ (\DAG) and a four-tuple  $\langle \U, \V, F, p(\U)\rangle$, where $\U$ is a set of \emph{exogenous} background variables distributed according to $p(\U)$, $\V$ is a set of observed \emph{endogenous} variables and $F = \{f_1, \dots, f_{|\V|}\}$ is a set of functions constituting the structural causal model (\scm) such that $v_i = f_i(pa_i, u_i)$ with $\text{pa}_i$ denoting the parents of $V_i$. Within $\V$, we distinguish between non-manipulative variables $\C$ that cannot be intervened on, \emph{continuous} treatment variables $\X$ that can be set to specific values and a single output variable $Y$ representing the agent's outcome of interest -- see \cref{fig:graph_health} for an example. Under the \emph{Markov Assumption}, each variable $V_i$ is conditionally independent of its non-descendants given its parents, such that the joint distribution factorises as: $p(\V \mid \mathcal{G}) = \prod_{V_j \in \V}p(V_j \mid \mathbf{Pa}_{j}^{\mathcal{G}}, \mathcal{G})$. Denote by $\mathcal{P}(\X)$ the power set of $\X$ giving \emph{all} possible interventions we can perform in the graph. The set $\XI \in \mathcal{P}(\X)$ represents one such intervention set with $\VNI = \mat{V} \setminus \XI$ being the corresponding set of non-intervened variables. We assume \emph{causal sufficiency} \citep{eleni2015} and \emph{perfect} interventions \citep{peters2017elements} so that, for any set $\XI$, the interventional distribution $p(\VNI \mid \DO{\XI}{\x_{I}})$ resulting from setting $\XI$ to a value $\x_{I}$ obeys the following \emph{truncated factorization}:
\begin{equation*}
    \label{eq:markov_fact_trun}
    p(\VNI \mid \DO{\XI}{\x_{I}}, \mathcal{G})  = \prod_{\onevarV \in \VNI}  p(\onevarV \mid \mathbf{Pa}_{j}^{\mathcal{G}}  ) \Bigr|_{\XI = \x_{I}} 
\end{equation*}
where the vertical line represents evaluation of the expression at $\XI = \x_{I}$. \\
\textbf{Notation} Note that, throughout, lowercase denotes \emph{realizations} of random variables (r.v.), while uppercase denotes the r.v. 
We denote collected data by $\dataI  = \{\{ ( \x_{I}^{(i)} , \valVNI^{(i)}) \}_{i = 1}^{N}\}_{\XI \in \setint}$ where the \emph{exploration set} (\setint) is $\setint = \mathcal{P}(\X)$ or $\setint \subseteq \mathcal{P}(\X)$ if only a subset of interventions can be implemented in the system (e.g. if $\mathcal{P}(\X)$ is too large). Here, $\valVNI^{(i)}$ represents one set of values of the \emph{non-intervened variables} $\VNI$ in a mutilated graph where $\XI$ is set to $\x_{I}^{(i)}$.
Further, $\valVNI^{(i)}$ includes both a value for the target variable $y^{(i)}$ and a value for the remaining variables, denoted by $\valVNINY^{(i)} = \valVNI^{(i)} \setminus y^{(i)}$. The dataset $\dataset$ includes all data, that is observational data ($\mathcal{D}^{O}$) which correspond to $\indexI = \varnothing$ and interventional data ($\mathcal{D}^I$). 
Every time we intervene in the system we collect $N > 1$ samples from each interventional distribution, but the framework equally handles $N=1$. See App. Section A for a table describing the full notation.
%

\textbf{Problem statement}
We consider the causal global optimization problem introduced by \citet{cbo} and generalize it to settings with an \emph{unknown} causal graph. Given $\dataI$, we seek to identify the interventional set \emph{and} corresponding values optimizing the causal effect in the \emph{true} causal graph $\graph$:%
\begin{align}\label{eq_objective}
    \XIstar, \nocolexpval^{\star} = \underset{\substack{\XI \in \mathcal{P}(\X) \\ \nocolexpval\in D\left(\XI\right)}}{\argmin } \mathbb{E}[Y \mid \DO{\XI}{\x_{I}}, \graph] \text{,}
\end{align}%
where $D\left(\XI\right)$ denotes the interventional domain of $\XI$ and the causal effect depends on the true graph $\graph$. Solving the problem in \cref{eq_objective} is challenging as evaluating $\mathbb{E}[Y \mid \DO{\XI}{\x_{I}}, \graph]$ requires intervening in the real system at a \emph{cost}, which we assume to be given by $\Co(\XI, \x_{I})$. \emph{While \cref{eq_objective} is the objective of \cbo, since the graph is unknown the problem becomes more challenging.}  

\textbf{Remark}: Notice that, when the graph is known, one can solve the problem in \cref{eq_objective} resorting to \cbo. \emph{When the graph is unknown, there is no existing unified solution for the problem.} This paper offers such a solution and shows how, accounting for graph uncertainty, one can match the performance of \cbo, which requires knowledge of the true graph, in terms of convergence speed to the optimum. 
\vspace{-10pt}
\section{Methodology}
\vspace{-5pt}
There are \emph{three} main ingredients to our solution to the problem in \cref{eq_objective}: a set of surrogate models for the causal effects on the target node (\cref{sec:inf_causal_effects}) accounting for different sources of uncertainty, a posterior over the graph structure (\cref{sec:inf_causal_graph}), computed exploiting observational and interventional data, and an information-theoretic acquisition function (\cref{sec:acquisition}) balancing the trade-off between optimization and structure learning. We now detail each ingredient and provide the pseudocode for \ceo in \cref{alg:example}. 
\vspace{-15pt}
\subsection{Inference on the causal effects}\label{sec:inf_causal_effects}
\textbf{Prior Surrogate Models}
For each set $\XI \in \setint$ we place a Gaussian process \citep[\gptext, ][]{williams2006gaussian} prior on $\fI(\x_{I}) = \mathbb{E}[Y \mid \DO{\XI}{\x_{I}}, \graph]$ and construct a prior mean and kernel function that incorporates our current belief about the graph together with the observational and interventional data. Specifically, we define $ \fI(\x_{I}) \sim \gp(\mI(\x_{I}), \kI(\x_{I}, \x_{I}'))$ with:
\begin{align}
    \label{eq_surrogate}
    &\mI(\x_{I}) = 
    \widehat{\mathbb{E}}[Y \mid \DO{\XI}{\x_{I}}] \qquad \kI(\x_{I}, \x_{I}') = k(\x_{I}, \x_{I}') + \widehat{\mathbb{V}}[Y \mid \DO{\XI}{\x_{I}}],
\end{align}
where $k(\cdot,\cdot)$ is a problem-dependent kernel function of choice. As in the surrogate model of \cbo, we set $m_{I}$ as an empirical estimate of the underlying true function using observational data. However, here we additionally incorporate the uncertainty over the graph by introducing latent variable $G$:
\begin{align}
     \widehat{\mathbb{E}}[Y \mid \DO{\XI}{\x_{I}}] &= \expectation{}{\hatexpectation{}{Y | \DO{\XI}{\x_{I}}, G }} \label{eq:prior_mean_int}
    \\
  \widehat{\mathbb{V}}[Y \mid \DO{\XI}{\x_{I}}] &=  \expectation{}{\hat{\mathbb{V}}[Y\mid \DO{\XI}{\x_{I}}, \graphvar]} + \mathbb{V}[\hatexpectation{}{Y\mid \DO{\XI}{\x_{I}}, \graphvar}] \label{eq:prior_var_int} \text{,} 
\end{align}
where outer expectations/variances are w.r.t a probability mass function on the r.v. $G$, i.e. $P(G)$; $\hat{\mathbb{E}}$ denotes that the expectation involves some approximation. Here, the approximation arises because we use $\hat{p}(Y\mid \DO{\XI}{\x_{I}}, \graphvar=g)$, an estimate of the interventional distribution computed via the do-calculus with only observational data \footnote{This can be computed when the causal effect is identifiable \citep{pearl2009causality}. See App. Section B for a discussion of the conditions under which \cref{eq:prior_mean_int} and (\ref{eq:prior_var_int}) converge to the true  $\mathbb{E}[Y \mid \DO{\XI}{\x_{I}}, \graph]$ and $\mathbb{V}[Y \mid \DO{\XI}{\x_{I}}, \graph]$ respectively.}.
Note that, when collecting data, $P(\graphvar)$ will be replaced by $P(\graphvar \mid \dataI)$, which also includes interventions. \emph{Thus, our models combine observational and interventional data}. Computing this posterior will be discussed in detail in \cref{sec:inf_causal_graph}.

\textbf{Surrogate Model Likelihood} For each intervention set $\XI \in \setint$ and value $\x_{I}^{(i)}$, we assume the output $y$ to be a \emph{noisy} realisation of the objective function $y = f_I(\x_{I}) + \nu$ at $\x_{I}^{(i)}$ where $\nu \sim \mathcal{N}(0, s^2)$. Indeed, every time we perform $\DO{\XI}{\x_{I}^{(i)}}$, we obtain a single sample $\mathbf{v}_{I}^{(i)}$ from the resulting interventional distribution $p(\mathbf{V}_I |\DO{\XI}{\x_{I}^{(i)}}, \graph)$. Noisy observations were not considered by \cbo, which made a simplifying assumption, and complicate the identification of the optimal intervention. 

\begin{figure*}[t!]
    \centering
    \begin{subfigure}[t]{0.33\textwidth}
        \centering
        \includegraphics[width=0.8\textwidth]{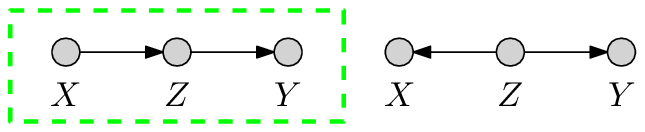}
        \includegraphics[width=1\textwidth]{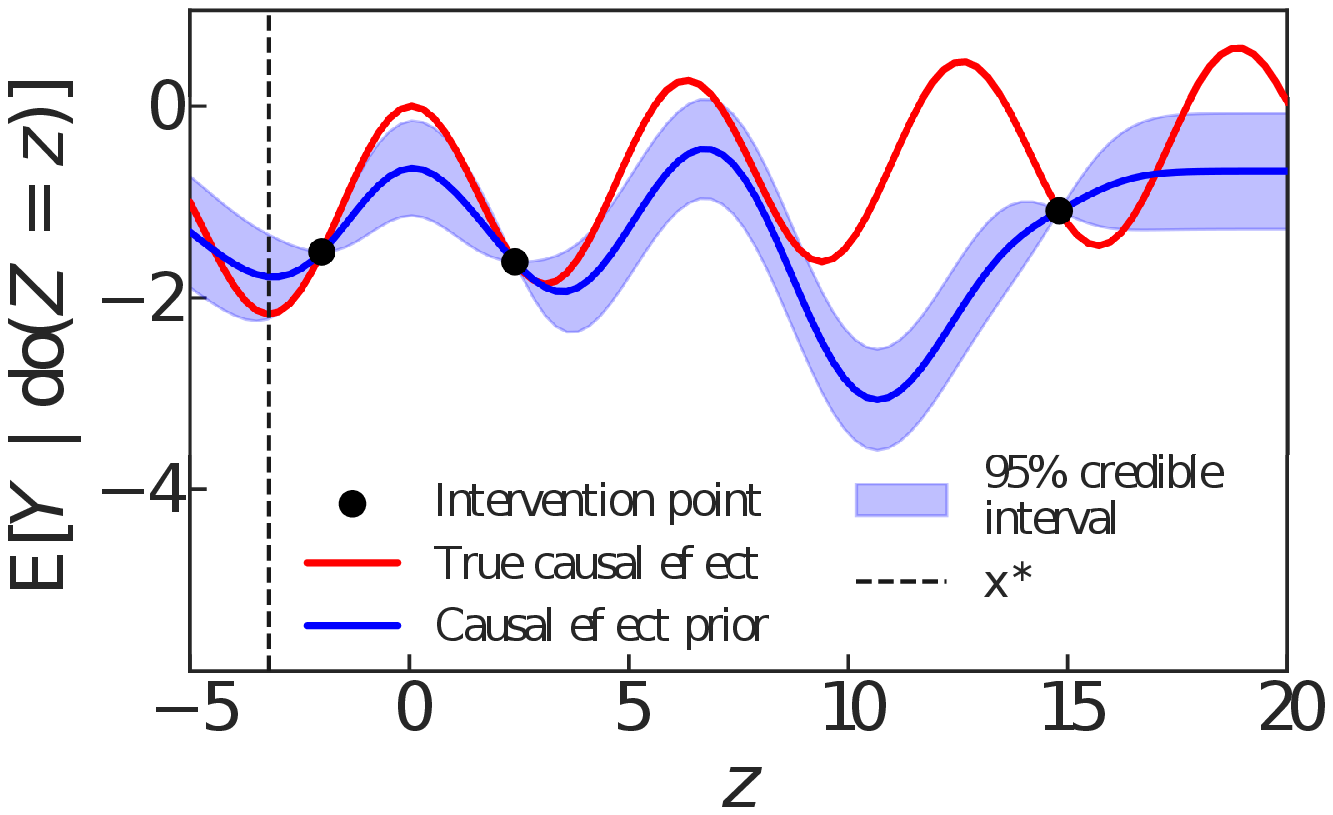}
        \caption{\label{fig:dag_prior_1}}
    \end{subfigure}%
    \hfill
    \begin{subfigure}[t]{0.33\textwidth}
        \centering
        \includegraphics[width=0.35\textwidth]{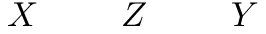}
        \includegraphics[width=0.93\textwidth]{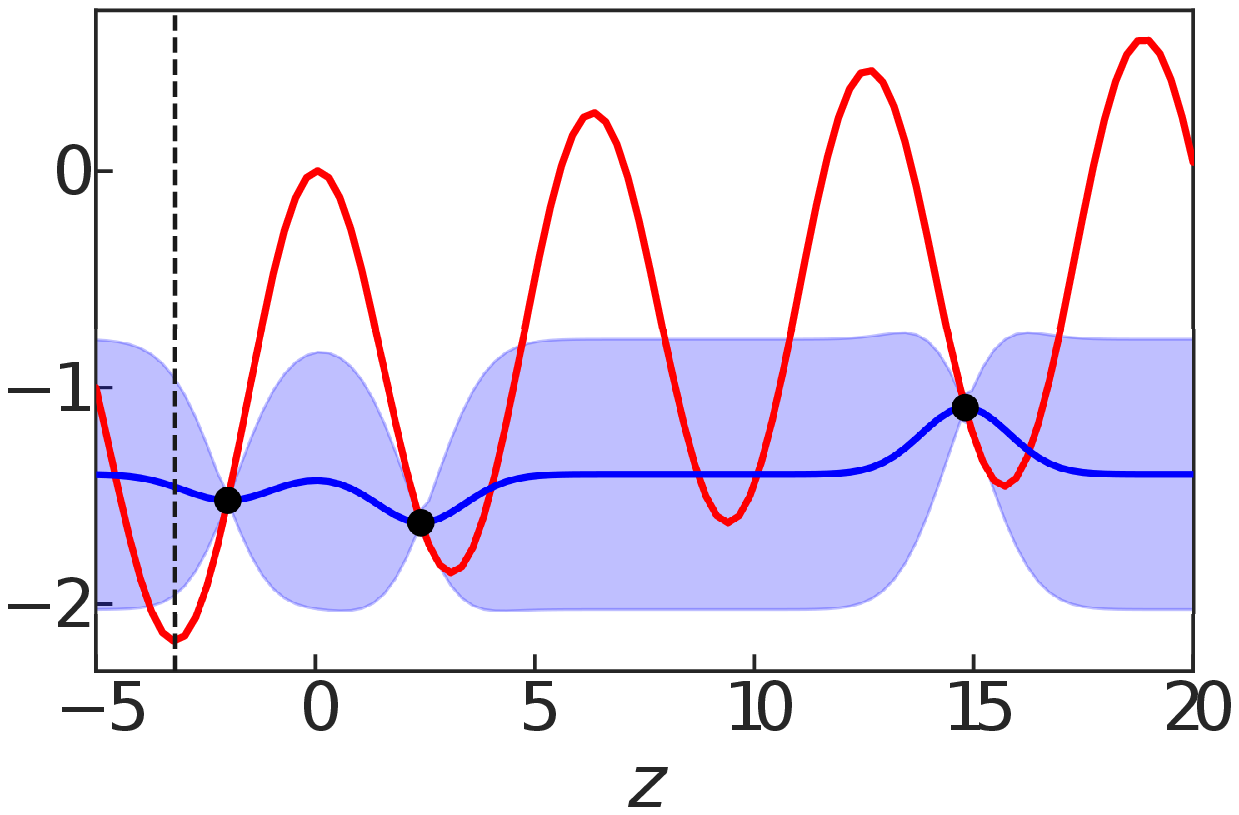}
        \caption{ \label{fig:dag_prior_2}}
    \end{subfigure}%
    \hfill
    \begin{subfigure}[t]{0.33\textwidth}
        \centering
        \includegraphics[width=0.35\textwidth]{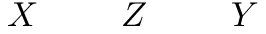}
        \includegraphics[width=0.93\textwidth]{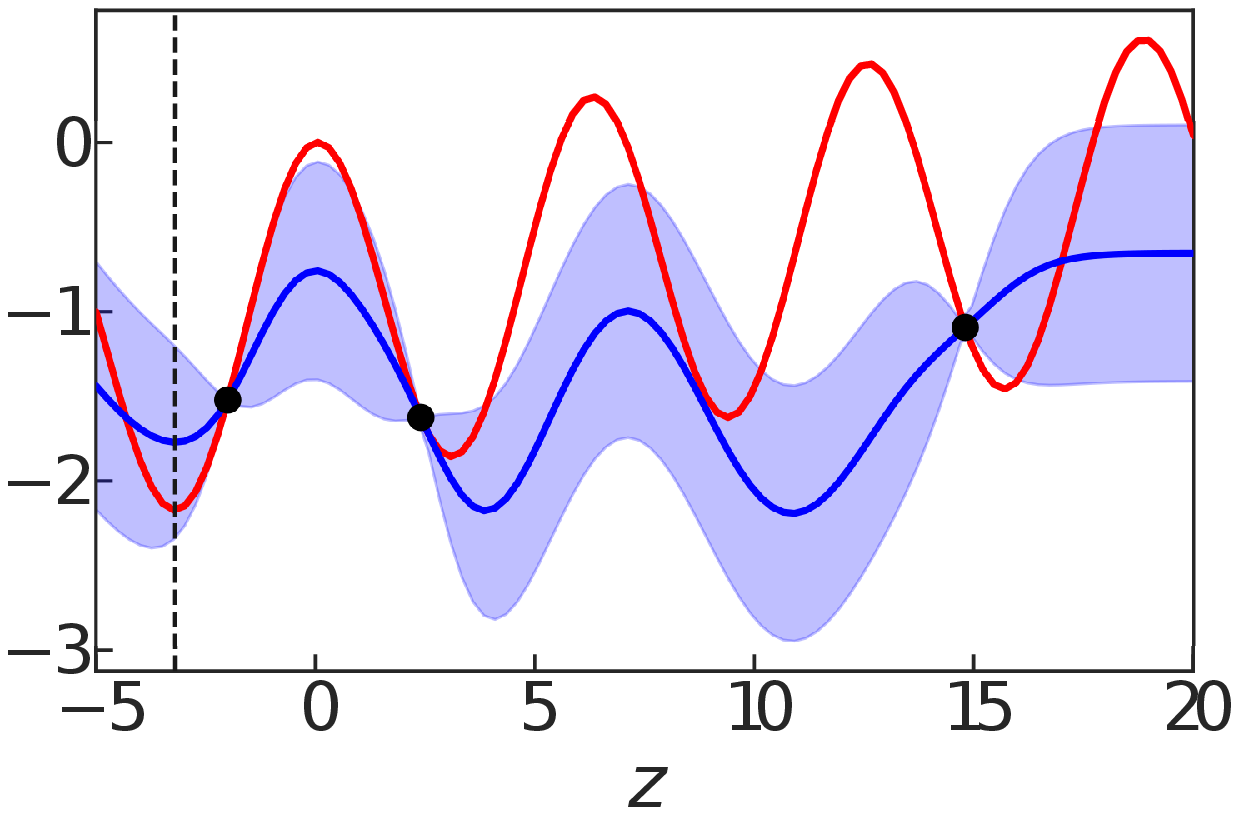}
        \caption{ \label{fig:dag_prior_3}}
    \end{subfigure}%
    \vskip -2pt
    \caption{Each figure plots the true underlying causal effect function $\mathbb{E}[Y \mid \DO{Z}{z},\graphvar = \graph]$ in red, while showing different surrogate models arising from \cref{eq:prior_mean_int} and \cref{eq:prior_var_int} when assuming each of the graphs of top of the plot is the correct one. The surrogate models share hyper-parameters so the differences only stems from alternative causal assumptions. The true \DAG $\graph$ is shown inside the dashed box in \cref{fig:dag_prior_1}. Vertical dashed line indicates the global optimum. Note that both the true graph and the second \DAG above (a) result in the same surrogate model.
    \label{fig:illustration}
    }
\end{figure*}


\textbf{Posterior Surrogate Models} Given the Gaussian likelihood, the posterior distribution $p(\fI \mid \dataset)$ can be derived analytically and will also be a \gptext with mean $\mI(\mathbf{X} \mid \dataset)$ and covariance functions $ \kI(\x_{I}, \x_{I}' \mid \dataset)$ computed by standard \gptext updates \citep[p. 19]{williams2006gaussian}.

\subsection{Inference on the causal graph}\label{sec:inf_causal_graph}
Given  $\dataset$, we update the prior distribution on $\graphvar$ to get $P(\graphvar \mid \dataset)$, so that we can account for the updated uncertainty in the surrogate models. 

We follow the setting of \citet{von2019optimal}, considering a discrete uniform prior distribution on $\graphvar$ with $P(\graphvar = g) = \frac{1}{|R_G|}$, and graphs can be enumerated.
This is appropriate for intended applications of \cbo, where a manageable number of graphs representing alternative causal hypotheses is maintained. However, we note that in our framework only expectations of the graph posterior are required. 

\textbf{Graph Likelihood}  In order to define the graph likelihood,we assume an additive noise model with \iid Gaussian noise terms\footnote{This is a common assumption, see \eg \citet{von2019optimal} or \citet{silva2010gaussian}.}. For every $g$ and $X_j \in \X$, we have:
\begin{align}
    X_j = \fj(\mathbf{Pa}_{j}^{\graphvar=g}) + \epsilon_j, \;\;  \fj \mid (\mathbf{Pa}_{j}^{\graphvar=g} = \x,\theta_j) \sim \mathcal{GP}(0, k_{j}(\x, \x^\prime; \theta_j))
    \label{eq:our_sem_eq} 
\end{align}
 with $\epsilon_j \sim \mathcal{N}(0, \sigma^{2}_{j})$ and where $\theta_j$ represents a set of hyper-parameters. Further, $\x$ and $\x^\prime$ are two values for the parents of $V_j$ in $G=g$ and $k_{j}$ is a kernel function of choice. Exploiting the available data we can then compute $p(\fj \mid \mathbf{Pa}_{j}^{\graphvar}, \mathcal{D}^{O})$ in closed form. Hence, each term in the truncated factorization is a \gptext marginal likelihood (details in App. Section B.)

 
\textbf{Graph Posterior} Given the prior distribution on $\graphvar$ and the likelihood, we can compute the posterior distribution on $\graphvar$ \emph{in closed form} as  $P(\graphvar \mid \dataI) \propto p(\dataI|\graphvar) P(\graphvar)$ due to the tractability of \gptext{s}. Importantly, this also allows us to compute the expectations w.r.t. $P(G|\mathcal{D})$ without approximations in both \cref{eq:prior_mean_int,eq:prior_var_int} and later in the acquisition function. Regarding convergence to $\delta_{G = \mathcal{G}}$, the key requirements needed are: causal sufficiency; that $R_G$ includes $\mathcal{G}$; all variables can be manipulated; infinite samples can be obtained from each node; causal minimality. See \cref{sec:graphposterior} for details.

\textbf{Effect of graph knowledge on optimization} It is important to emphasize that while computing $P(\graphvar \mid \dataI)$ and using it in the surrogate model provides proper uncertainty quantification, exact knowledge of the causal structure is not \emph{always} needed for more efficient causal optimization.

Indeed, different graphs may lead to equivalent surrogate models. For instance, consider the example in \cref{fig:illustration} showing different causal graphs and the associated surrogate models, when three interventions are collected. Here the true \DAG is given in \cref{fig:dag_prior_1} (green dashed box) alongside an alternate causal \DAG. The surrogate models for these structures are equivalent, and shown in the plot below. Indeed, the do-calculus for the left \DAG and right \DAG of \cref{fig:dag_prior_1} gives the same result, even if the latter \DAG is wrong. Given our surrogate model construction in (\cref{eq_surrogate}), this implies the same prior mean and covariance for the causal effects associated with both graphs. Therefore, for the purpose of causal optimization, the \DAG{s} in \cref{fig:dag_prior_1} are the same.

As demonstrated experimentally in \cref{sec:experiments}, \emph{it is thus wasteful to intervene to find the true graph first, and only after perform causal optimization}. This motivates our joint approach, which automatically balances structure learning and optimization by picking interventions that are reducing the uncertainty of $P(\graphvar \mid \dataI)$ \emph{when} doing so enables faster identification of the optimum. Next, we describe how to achieve this balance by proposing a new acquisition function.

\subsection{Acquisition function: Causal Entropy Search} \label{sec:acquisition}
We propose an acquisition function that combines Bayesian structure learning and causal optimization into a single objective in order to balance the two tasks. To do so, we start by framing the tasks of causal optimization and structure learning, respectively, as \emph{gaining information} about the global optimum \emph{value} $y^\star =  \min (\text{or} \max )_{I}  y^{\star}_I $ and about $G$. This information gain can be quantified by considering the reduction in uncertainty in an appropriately defined joint distribution on $y^\star$ and $G$. While a distribution on the graph, $P(G\mid \dataI)$, was defined in \cref{sec:inf_causal_graph}, the distribution on $y^\star$ is implicitly \emph{induced} by the surrogate models defined in \cref{sec:inf_causal_effects}. \footnote{This is because the distribution on each $y^{\star}_I$, induced by the prior on $f_I$, implies a distribution on the global optimum value achieved $y^\star$.} 

 Our acquisition is thereby defined as the conditional mutual information (\mi) denoted by $\mathbb{I}(\cdot;\cdot\mid\cdot)$ between the random variables $(y^\star, G)$ and the \emph{outcome of the experiment} $\left(\mathbf{v}_{Y}, y \right)$, given data $\dataI$. An experiment in the context of a causal \DAG consists of performing $\DO{\XI}{\x_{I}}$ and observing a sample $\valVNI = (y, \mathbf{v}_Y)$ from the resulting interventional distribution. Therefore, in \ceo we seek an intervention set $\XI^{\text{best}}$ \emph{and} corresponding value $\x_{I}^{\text{best}}$ such that:
  \begin{align}
        \X_I^{\text{best}}, \x_{I}^{\text{best}} &= \argmax_{\XI \in \setint ,~\nocolexpval\in D\left(\XI\right)} \alpha_{\our}(\mathbf{X}_{I} , \mathbf{x}_I)\nonumber \\
      \alpha_{\our}(\mathbf{X}_{I}, \mathbf{x}_I) &= \frac{\mathbb{I}\left[\left(y^{\star}, G\right) ;\left(\mathbf{X}_{I}, \mathbf{x}_{I}, \mathbf{v}_{Y}, y \right) \mid \dataI \right]}{\Co(\XI, \x_{I})} \label{eq:ces_acq}
  \end{align}
where the denominator $\Co(\XI, \x_{I})$ provides the \mi per unit of cost. We call the acquisition function in \cref{eq:ces_acq} Causal Entropy Search (\ces). 

Assume, for the moment, that we have access to $p(y^\star, G)$ and the associated posterior. We write its joint entropy as: 
\begin{align}\label{eq:joint_entropy}
    &\mathbb{H}[p(y^\star,G \mid \dataset)] = -\sum_G \int \mathrm{d} y^\star p(G,y^\star \mid \dataset )  \log p(G,y^\star \mid \dataset) \text{.} 
\end{align}
The work in \citep[Lemma 2.3 and Eq. 2.2]{marx2021estimating} shows that conditional MI and joint entropies can be rigorously defined in an analogue way to their fully continuous/discrete counterparts. The book \citep{cover1999elements} gives conditions on densities required for the differential entropy to exist and be bounded. We can then use \cref{eq:joint_entropy} to evaluate the numerator of \ces, i.e. \cref{eq:ces_acq}, which can be written as:
\begin{align*}
     \mathbb{E}_{p(\mathbf{v}_{Y}, y \mid \DO{\XI}{\nocolexpval}, \dataI)  } \Big[ \mathbb{H}[p(y^\star,G \mid \dataI)] -   \mathbb{H}[p(y^\star,G \mid \dataI \cup \left(\mathbf{x}_{I}, \mathbf{v}_{Y}, y \right) )]  \Big] \text{.}
\end{align*}
This is the reduction in entropy observed \emph{on average} in the joint distribution on $(y^\star,G)$ when performing $\DO{\XI}{\nocolexpval}$ and observing $\mathbf{v}_{Y} = (\mathbf{v}_{Y}, y)$. We note that computing this term does not require intervening in the real system. For a given intervention we ``fantasize'' (a concept often used in \bo \citep{wilson2018maximizing}) about the outcomes $\mathbf{v}_{Y}$ we would get by simulating what would happen in the system under $\DO{\XI}{\nocolexpval}$. This can be done in \ceo by using the surrogate models and the fitted \scm functions. Note that this formulation automatically takes into account noisy observations, which often motivates entropy-based acquisitions in \bo \citep{frazier2018bayesian}. We now discuss how to define and obtain $p(y^\star, G)$, and approximate $\mathbb{H}[p(y^\star,G)]$.
\begin{figure}
\begin{minipage}[t]{0.5\linewidth}
        \begin{table}[H]
        \vspace{-0.5cm}
            \caption{\begin{small} Causal Entropy Search compared to related objectives in \boed for structure learning and \bo algorithms. Entropy Search (\acro{es}) and Max-value \acro{es} are used within the standard a-causal \bo problem. Multi-Task \bo is also used for a-causal optimization with multiple functions, but when the function containing the optimum is known. The objective in \citep{von2019optimal} is also based on \mi, but only targets the graph structure.  \ces considers intervention sets, intervention values and the graph structure when the function containing the optimum is unknown. \vspace{0.5em}\end{small}}
        \begin{tabularx}{\textwidth}{XX}
        \toprule
        \textbf{Acquisition} & \textbf{Objective} \\
        \midrule
                \textbf{\ces (ours)}  & $\frac{\mathbb{I}\left[\left(y^{\star}, G\right) ;\left(\mathbf{X}_{I}, \mathbf{x}_{I}, \mathbf{v}_{Y}, y \right) \mid \dataI \right]}{\Co(\XI, \x_{I})}$  \\[10pt]
                Entropy Search  \citep{hennig2012entropy}    &   $\mathbb{I}\left [ (\mathbf{x}, y); \mathbf{x}^\star  \right | \mathcal{D} ]$  \\[10pt]
                Max-value Entropy Search \citep{wang2017permutation}    &   $\mathbb{I}\left [ (\mathbf{x}, y); y^\star  \right | \mathcal{D} ]$  \\[10pt]
                Multi-Task \bo  \citep{NIPS2013_f33ba15e}    &   $\frac{\mathbb{I}\left [ (\mathbf{x}^{\text{task}}, y); \mathbf{x}^\star  \right | \mathcal{D} ]}{ \Co(\text{task})}$ \\[10pt]
                Causal discovery via \bo  \citep{von2019optimal} & $\mathbb{I}\left[G ;\left(\mathbf{X}_{I}, \mathbf{x}_{I}, \mathbf{v}_{I}\right) \mid \dataI \right]$ \\
                \bottomrule
            \end{tabularx}
            \label{table:acquisitions}
        \end{table}
\end{minipage}%
\hfill
\noindent
\begin{minipage}[t]{0.46\linewidth}
    \raggedleft
        \begin{algorithm}[H]
           \caption{\our}
           \label{alg:example}
            \begin{algorithmic}[1]
               \STATE {\bfseries Input:} $\dataI$, $P(G)$, $H$ (N. of iterations)
               \STATE {\bfseries Output:} $\XIstar, \x_{I}^{\star}, P(G \mid \mathcal{D}_H)$.
               \STATE {\bfseries Initialise:} Set $\dataI_0 = \dataI$.
               \STATE Compute $p(\dataI_0 \mid \graphvar = g)$ for each $g \in \suppG$. 
               \STATE Compute $P(\graphvar \mid \dataI_0)$.
               \STATE Compute $\mI(\x_{I})$ and $\kI(\x_{I}, \x_{I}')$ using the do-calculus for  each $\XI \in \setint$  (\cref{eq:prior_mean_int} and \cref{eq:prior_var_int}).
               \FOR{$h=1, ..., H$}
               \STATE Compute $ \alpha_{\our}(\mathbf{X}_{I}, \mathbf{x}_I)$  for each $\XI \in \setint$  and each $\mathbf{x}_I$ in the acquisition points with \ces (Algorithm 2 in App.).
               \STATE Obtain the optimal set-value pair $(\XI^h, \x_{I}^h)$. 
               \STATE Intervene on the system and augment the dataset $\dataI_h = \dataI_{h-1} \cup (\x_{I}^h, \valVNI^{h})$.
               \STATE Compute $P(G \mid \dataI_h)$ and every $p(\fI \mid \dataI_h)$.
               \ENDFOR
            \STATE Return $(\mathbf{X}_{I}^\star, \mathbf{x}_{I}^\star), P(G | \mathcal{D}_H)$

            \end{algorithmic}
        \end{algorithm}
\end{minipage}
\vspace{-0.3cm}
\end{figure}
 
\textbf{Joint posterior over $y^\star$ and $G~~$} 
Computing \ces requires defining $p(y^\star, G)$, obtaining its posterior given $\mathcal{D}$ and computing \cref{eq:joint_entropy}. We model the joint distribution as $p(y^\star, G \mid \dataI) = P(G \mid y^\star, \dataI) p(y^\star| \dataI)$\footnote{The alternative model $p(y^\star | G, \dataset ) P(G | \dataset)$ could also be considered. We discuss this in App. Section F}. We can thus write \cref{eq:joint_entropy} as the sum of:
$-\sum_{G}\int \mathrm{d} y^\star P(G\mid y^\star , \dataset) p(y^\star \mid \dataset) \log P(G \mid y^\star, \dataset)$
and $\mathbb{H}(p(y^\star \mid \dataset))$. In turn, evaluating these two terms require computing $P(G \mid y^\star, \dataset )$ and sampling from $p(y^\star \mid \dataset)$. The former can be achieved by updating the posterior on $\graphvar$ to get $P(G \mid \dataset \cup ( \x^\star, \mathbf{v}_{Y}^{\star}, y^\star))$ and marginalizing over $ \x^\star$ and $\mathbf{v}_{Y}^{\star}$ (see App. Section C.1 for details). 

However, since $p(y^\star \mid \dataset)$ is not in available closed form, and a simple Monte Carlo approximation is computationally expensive, we propose a heuristic approximation that works in practice. Specifically, we approximate the distribution of $y^\star$ to be a \emph{mixture} of the intervention-specific distributions on the optimal targets, where the weights are given by the probability of each intervention set being optimal:
\begin{align}\label{eq:pystar}
       p(y^\star \mid \dataset) = \sum_{y_{I}^{\star} : \mathbf{X}_I \in \setint} P(\mathbf{X}_I = \mathbf{X}^{\star}) ~p(y_{I}^{\star} \mid  \dataI )  ,
\end{align}
where $\mathbf{X}^{\star}$ is the intervention set in $\setint$ which yields the global optimum. We can thus sample from \cref{eq:pystar} via standard mixture sampling techniques \citep{mcbook}.
 Notice that \cref{eq:pystar} is accounting for the fact that we do not know which intervention set and thus surrogate model is associated with the global optimum\footnote{Therefore, this formulation provides a useful acquisition that could be used not only for causal optimization, but also in a-causal settings (see App. Section H for details).}.
See \cref{table:acquisitions} for connections between the \ces and related existing objectives.\\
\textbf{Motivation of posterior approximation} Notice that as we collect data and get closer to the optimum, we expect the weights of the mixture distribution \cref{eq:pystar} gradually concentrate around the optimal intervention set $\mathbf{X}^{\star}$, and $p(y^\star \mid \dataset)$ turns into $ p(y_{I}^{\star} \mid \dataset)$ for the $\XI$ s.t. $\XI =\mathbf{X}^{\star}$. To model the belief over the optimal set $P(\mathbf{X}_I = \mathbf{X}^{\star})$, we employ a \emph{multi-armed bandit} \citep{lattimore2020bandit} perspective and define the weights via an upper confidence bound (UCB)  policy -- see App. Section H for details.\\
\textbf{Computational complexity}
Let $N$ be the number of \emph{acquisition points} i.e., given $\mathbf{X}_I$, we will consider $\{ \mathbf{x}_{I}^{(n)} \}_{n=1}^{N}$ as candidate values to compare with \ces. The total complexity of \ces can be written: $\mathcal{O}(N \cdot \text{CES}(\mathbf{x}) \cdot \sum_{\mathbf{X}_I \in \mathbf{ES}} |\mathbf{X}_I|  )$. Here, $\text{CES}(\mathbf{x})$ denotes the time needed to compute \ces for m specific value $\mathbf{x}$, regardless of its corresponding $\mathbf{X}_I$. Note this is valid for CBO also, replacing $\text{CES}(\mathbf{x})$ with $\text{CEI}(\mathbf{x})$, the acquisition used by CBO \citep{cbo}. Here, $\text{CES}(\mathbf{x})$ involves approximating univariate marginals $p(y_{I}^{\star} \mid  \dataI )$ and the mixture \cref{eq:pystar}. We do this with Kernel Density Estimation (KDE); the complexity of these operations depend on the accuracy needed to estimate these marginals. Finally, 
since the number of graphs in our setting is tractable, we can compute expectations w.r.t $P(G)$ without approximations and with negligible cost w.r.t to the rest.\footnote{For more details, see App. Section H. }
\vspace{-5pt}
\section{Related work}
\textbf{Causal Effect Estimation} A variety of approaches to estimate causal effects from observational data have been proposed, including those based on propensity scores \citep{rosenbaum1983central}, instrumental variables \citep{angrist1995identification} or \scm{s} \citep{pearl2009causality}. On the contrary, there have been  few methods combining interventional and observational data \citep{silva2016observational}. Focusing on \scm methods, apart from a few exceptions \citep{hyttinen2015calculus, horii2021bayesian}, causal effects are estimated assuming \emph{exact} knowledge of $\mathcal{G}$.

\textbf{Causal Discovery} The majority of causal discovery (\cd) methods focus on learning $\mathcal{G}$ using only observational data thus restricting the identification to the Markov equivalence class (\mec) \citep{verma1991equivalence, andersson1997characterization, spirtes2000constructing, chickering2002optimal,friedman2003being, shimizu2006linear, janzing2012information, zhang2015estimation}. 
The seminal work by \citet{cooper_yoo99} first showed how experimental design can improve causal structure learning (which in general is known to be \emph{NP-hard} \citep{chickering1996learning}). Since this study, several papers have focused on learning $\mathcal{G}$ from a combination of interventional \emph{and} observational data \citep{tong2001active,murphy2001active,eaton2007exact,hauser2012characterization, hauser2014two, hauser2015jointly, wang2017permutation, ness2017bayesian, yang2018characterizing, ghassami2018budgeted, abcd, faria2022differentiable}. However, all of these focus on finding the true graph by selecting the intervention \emph{set} only. Our work additionally selects intervention \emph{values}.
Recently, \citet{von2019optimal} developed a \bo framework for \cd where variables are continuous and follow flexible non-linear relationships. 

\textbf{Optimal Causal Decision Making} The literature on causal decision making has mainly focused on finding the optimal treatment regime using observational data \citep{zhang2012robust, atan2018deep, haakansson2020learning}. The idea of identifying the optimal action or policy by performing interventions in a causal system has been explored in causal bandits \citep{lattimore2016causal}, causal reinforcement learning \cite{zhang2020designing} and, more recently, in \bo \citep{cbo, daggp, aglietti2021dynamic}. Importantly, all these approaches assume exact knowledge of the causal relationships beforehand, an assumption that is often not met in practice. Recent work attempts to relax this for causal bandits \citep{lu2021causal,wang2021actively}.

\textbf{Entropy in Causality}
The use of information-theoretic concepts such as entropy in a causal setting has been receiving increased interest. \citet{kocaoglu2017entropic} introduces the framework ``Entropic Causal Inference', which exploits an entropy condition on a exogenous variable to find the causal direction between two categorical nodes from observational data (extended in \citep{compton2022entropic}); a related notion, ``common entropy'', for two discrete random variables is discussed in \citep{kocaoglu2020applications} which 
improves classic causal discovery algorithms such as PC and FCI. The works \citep{steudel2015information,miklin2017entropic} discuss the use of entropy among more than two variables. 
%
\vspace{-5pt}
\section{Experiments}
\label{sec:experiments}
We demonstrate \ceo on a benchmark synthetic example used in \citep{cbo} as well as real-world applications for which a \DAG is available and can be used as a simulator. Without loss of generality, we will always \emph{minimize} causal effects rather than maximize. Results are averaged over 12 replicates of different initial $\dataset^{I}$, while $\dataset^{O}$ is fixed. We set $|\dataset^{I}_0|=2$ and $|\dataset^{O}_0|=200$ (initial data) unless otherwise specified\footnote{A Python implementation will be released post-review.}. Full experimental details including \scm details, kernel functions used, hyper-parameter optimization can be found in Section K of the supplement. As a reminder for key notation, recall $\mathcal{G}$ denotes the true causal graph; $y^\star$ denotes the best value of the causal effect across $\mathbf{ES}$. 

\textbf{Experiments Roadmap \& Baselines} We compare the performance of \ceo against \cbo which is the state-of-the-art algorithm addressing the causal global optimization problem. However, as \cbo assumes knowledge of the graph, we run it first assuming the true graph $\graph$ and then assuming each of the wrong graphs. In general, \cbo with $\graph$ is expected to perform better, especially when the graph structure is particularly informative for the \cbo prior. However, this need not be the case in all experiments. Indeed, in one of our real examples, \cbo equipped with $\graph$ performs worse than \ceo. To further highlight the benefit of our joint method, in \cref{sec:comparison} we compare against an algorithm that first identifies the causal graph by optimizing the \mi\footnote{We note also note that \citep{abcd} use a \mi criteria for causal discovery via experimentation. However, this work only selects intervention set and not values and it is not not applicable to our continuous variables setting.} as considered by \citet{von2019optimal}, and then runs \cbo to select the optimal action retaining the interventional data collected when learning about the graph. This is to show that jointly considering structure learning and causal optimization leads to better performance than a sequential approach. We refer to this method as \acro{cd-cbo}. Notice that we cannot compare directly to \citet{von2019optimal} as in their work the \mi is optimized via \bo rather than computed as in \ceo. Finally, notice also that causal discovery methods which do not (1) collect interventions with active learning (2) select both intervention sets and values and (3) assume continuous variables, are not applicable to the problem setting thus we do not consider them as an alternative to \citet{von2019optimal} for \acro{cd-cbo}.

\textbf{Performance measures}
\begin{table}[t!]
\centering
\caption{Average \acro{gap} $\pm$ one standard error computed across 12 replicates initialized with different $\dataI$. Higher values are better, and $0 \leq$ \acro{gap} $\leq 1$. The best result for each experiment in bold. \\}
\label{tab:table_gaps}
\begin{tabular}{lcccc}
\toprule
Method & Synthetic (\cref{fig:dag_prior_1}) & Health (\cref{fig:graph_health}) & Epi (\cref{fig:epi_DAG}) & $\text{Epi}^{+}$ (Fig. 9 App. J) \\
\midrule
$\our$ & $\textbf{0.75} \pm 0.03$ & $0.62 \pm 0.07$ & $\textbf{0.59} \pm 0.06$ & $\textbf{0.65} \pm \textbf{0.03}$ \\
$\cbo$ w. $G = \graph$ & $0.66 \pm 0.04$  &  $\textbf{0.71} \pm 0.05$ &  $0.50 \pm 0.05$  & $0.50 \pm 0.03$\\
$\cbo$ w. $G \neq \mathcal{G}$ & $0.59 \pm 0.03$  & $0.62 \pm 0.06$ & $0.50 \pm 0.06$ & $0.53 \pm 0.03$\\
\bottomrule
\end{tabular}
\end{table}
We evaluate performance by assessing the convergence speed to the optimum value of $Y$ as measured by the total \emph{cumulative cost} of interventions taken 
where the cost of a single intervention is given by the number of variables in the intervened set. Further, we also evaluate our approach using the \acro{gap} metric introduced in \citet[Eq. (4)]{aglietti2021dynamic}.
Note that $0 \leq \acro{gap} \leq 1$ and that higher values are better.
\subsection{Synthetic example}
We start by considering data generated by the chain graph $X \rightarrow Z \rightarrow Y$ as per \cref{fig:dag_prior_1} (green dashed box). We consider \emph{all} possible alternative \DAG{s} with three nodes as alternative causal hypotheses, except for those which do not make sense for causal optimization: graphs with any isolated nodes or where the target $Y$ is not a sink. See App. Section I for the full list of graphs and the true \scm.
 
Convergence results are given in \cref{fig:convergence_results}, while we show the evolution of the posterior on the true graph over iterations in App. Section D. Note how \cbo run under incorrect causal assumptions does not converge to the global optimum, whereas \ceo matches the performance of \cbo run with the true graph even if the graph posterior has not converged. This is confirmed by the \acro{gap} values in \cref{tab:table_gaps} where \ceo outperforms both \cbo algorithms across 12 replicates.
  
\subsection{Comparison with learning the structure first, then performing \cbo (\acro{cd-cbo})}\label{sec:comparison}
We now study how \ceo compares against an algorithm that first performs causal discovery and, once the graph has been identified, solves the optimization problem via \cbo. We refer to this as \acro{cd-cbo}. We consider the graph learned once it has more than $90\%$ of posterior mass. In \cref{fig:convergence_results}(a) , notice the significantly worse performance at optimization of \acro{cd-cbo}. This is due to the fact that, while the \mi correctly identifies $\graph$ after a few samples for most graphs, the graph in \cref{fig:dag_prior_3} is very hard to distinguish from $\graph$ (for the given \scm) as the terms in the truncated factorization give similar likelihood values. Therefore, the posterior never gets concentrated around a single graph despite the high number of selected interventions, and the optimization task is never solved.  
This reflects the benefit of a joint approach, where the graph is learned along optimization of the effect, and only to the extent to which it is useful for optimization. Additional results are presented in \cref{sec:additionalresults}.
\subsection{Real examples}
We now study the performance of \ceo with causal \DAG{s} used in two real-world settings: one in healthcare to model the level of Prostate Specific Antigen (\psa) and one in epidemiology to model the level of \acro{hiv} virus load. These graphs exhibit a significantly more complex dependency structure than the chain graph of the synthetic example. For the graphs considered here, we found our posterior to converge immediately as soon as initial data $\dataset$ is provided, or after one/two interventions. \ceo can take advantage of this, and simply optimize the \mi for $y^\star$.

\textbf{Healthcare}
The \scm for this real-world application results in causal effects that are linear functions of their inputs, observed with standard white Gaussian noise. We designed four incorrect graphs which represent plausible hypotheses a doctor may have in this context (App. Section K).
Due to the simplicity of the underlying true functions, one can see that the greediness of \ei (used by \cbo) grants it better performance on average. \ceo still consistently outperforms \cbo on the wrong graphs. This also shows evidence that while exact knowledge of the graph is not always required for efficient optimization, it can be better on average than incorrect knowledge of it.

\begin{figure*}[t]
    \centering
    \begin{subfigure}[t]{0.25\textwidth}
        \centering
        \includegraphics[width=1\textwidth]{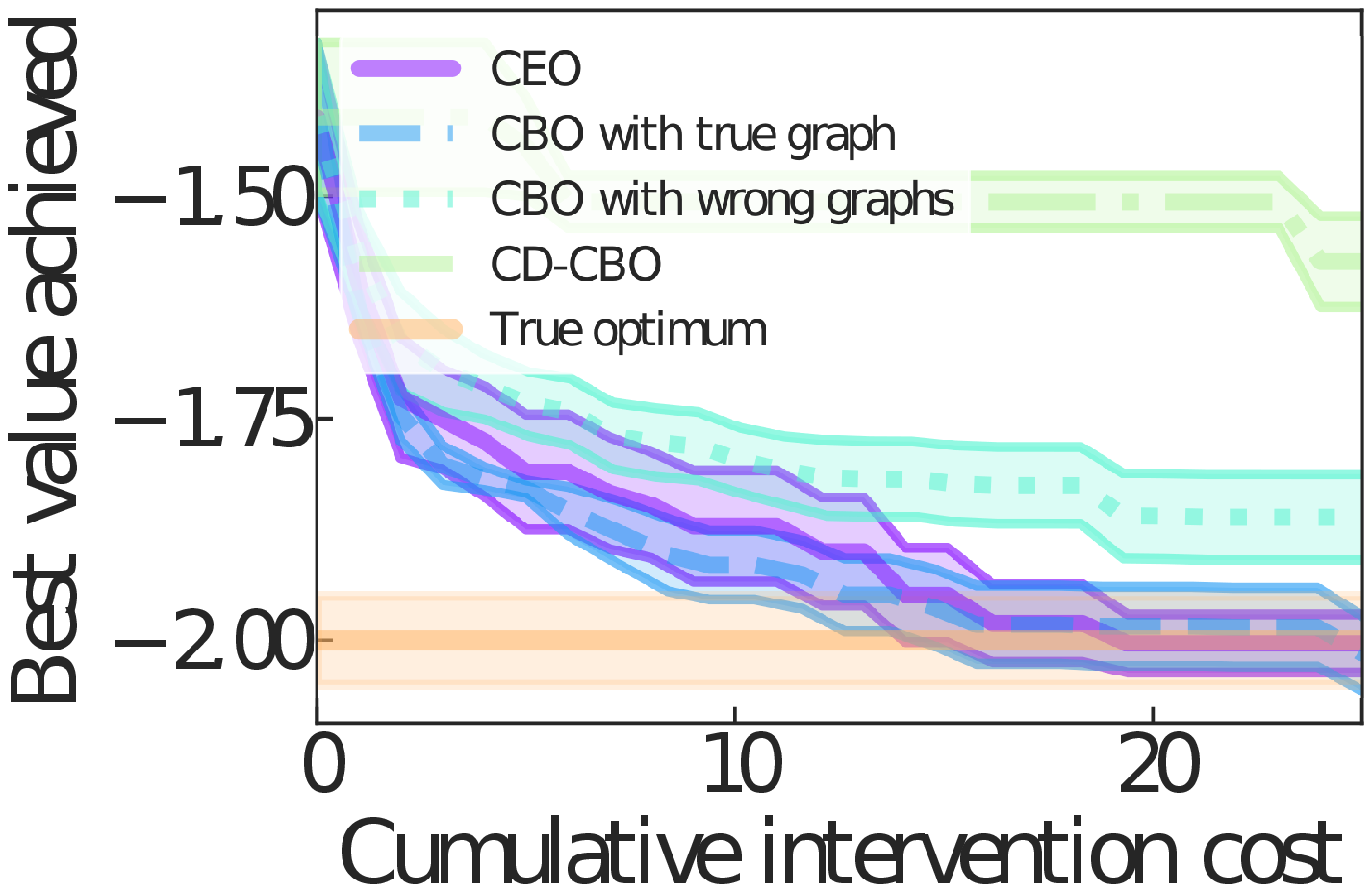}
        \caption{Synthetic.}
    \end{subfigure}%
    \hfill
    \begin{subfigure}[t]{0.24\textwidth}
        \centering
        \includegraphics[width=1\textwidth]{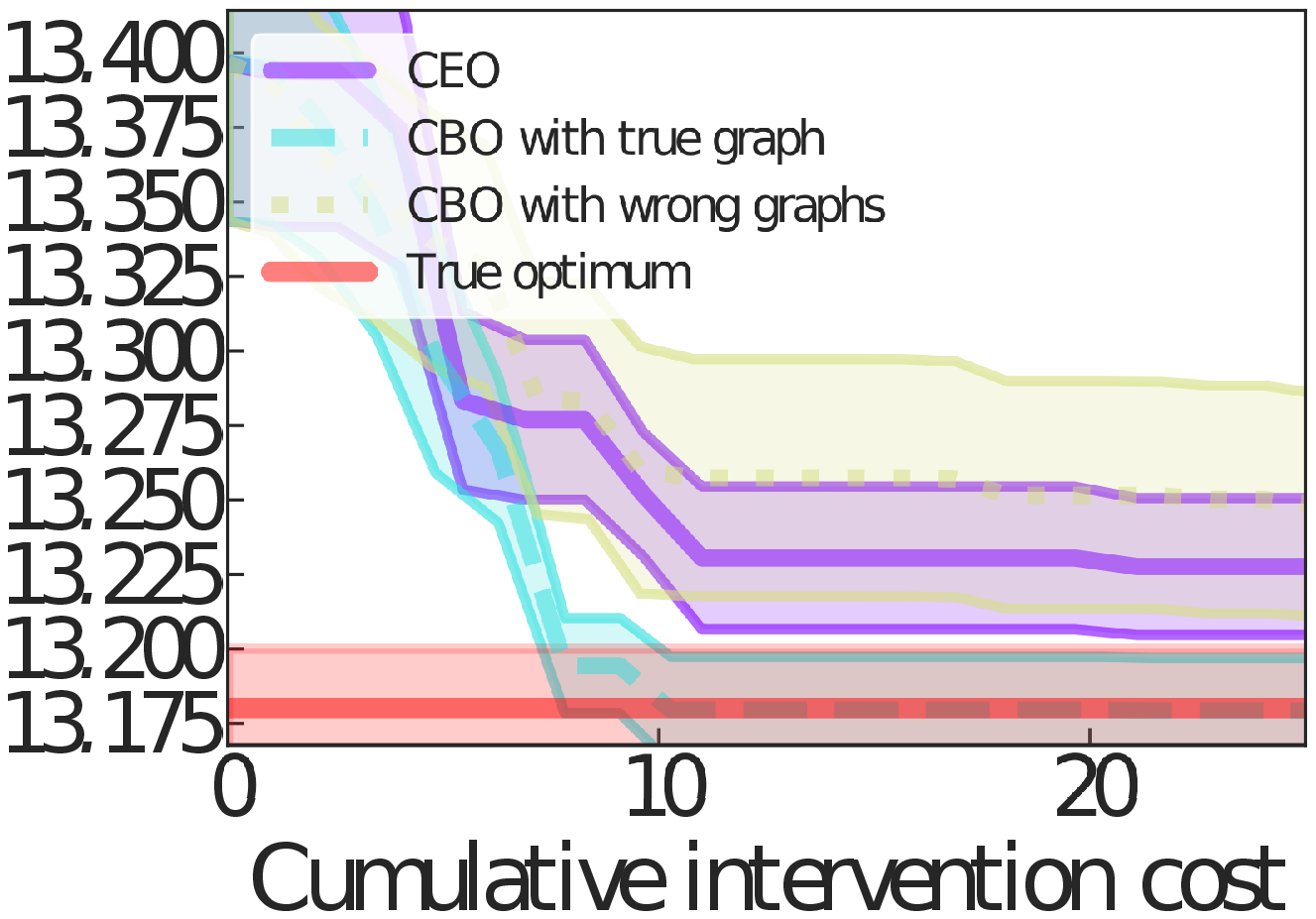}
        \caption{Healthcare.}
    \end{subfigure}%
    \hfill
    \begin{subfigure}[t]{0.24\textwidth}
        \centering 
        \includegraphics[width=1\textwidth]{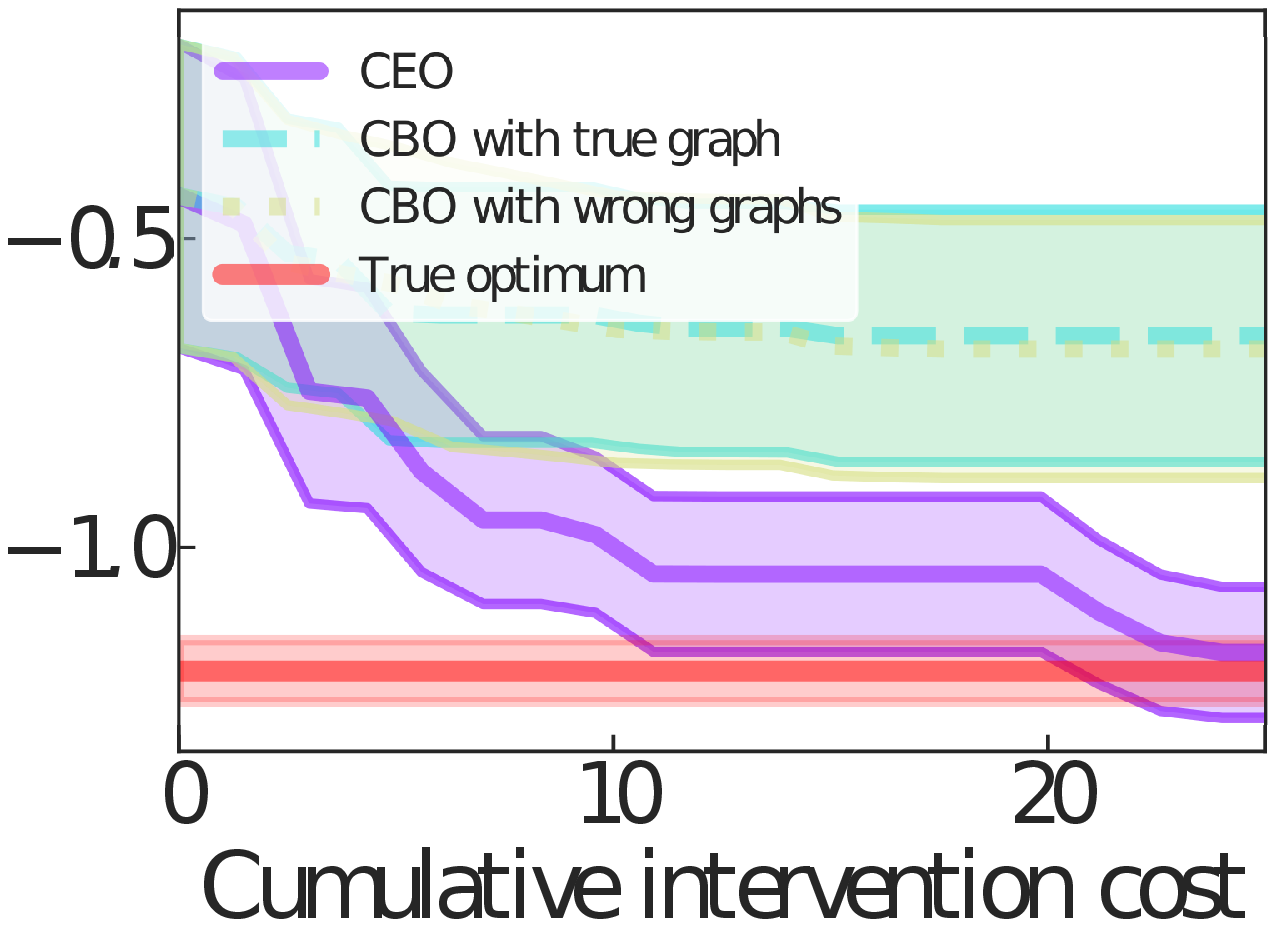}
        \caption{Epidemiology.}
    \end{subfigure}%
    \hfill
    \begin{subfigure}[t]{0.24\textwidth}
        \centering 
        \includegraphics[width=1\textwidth]{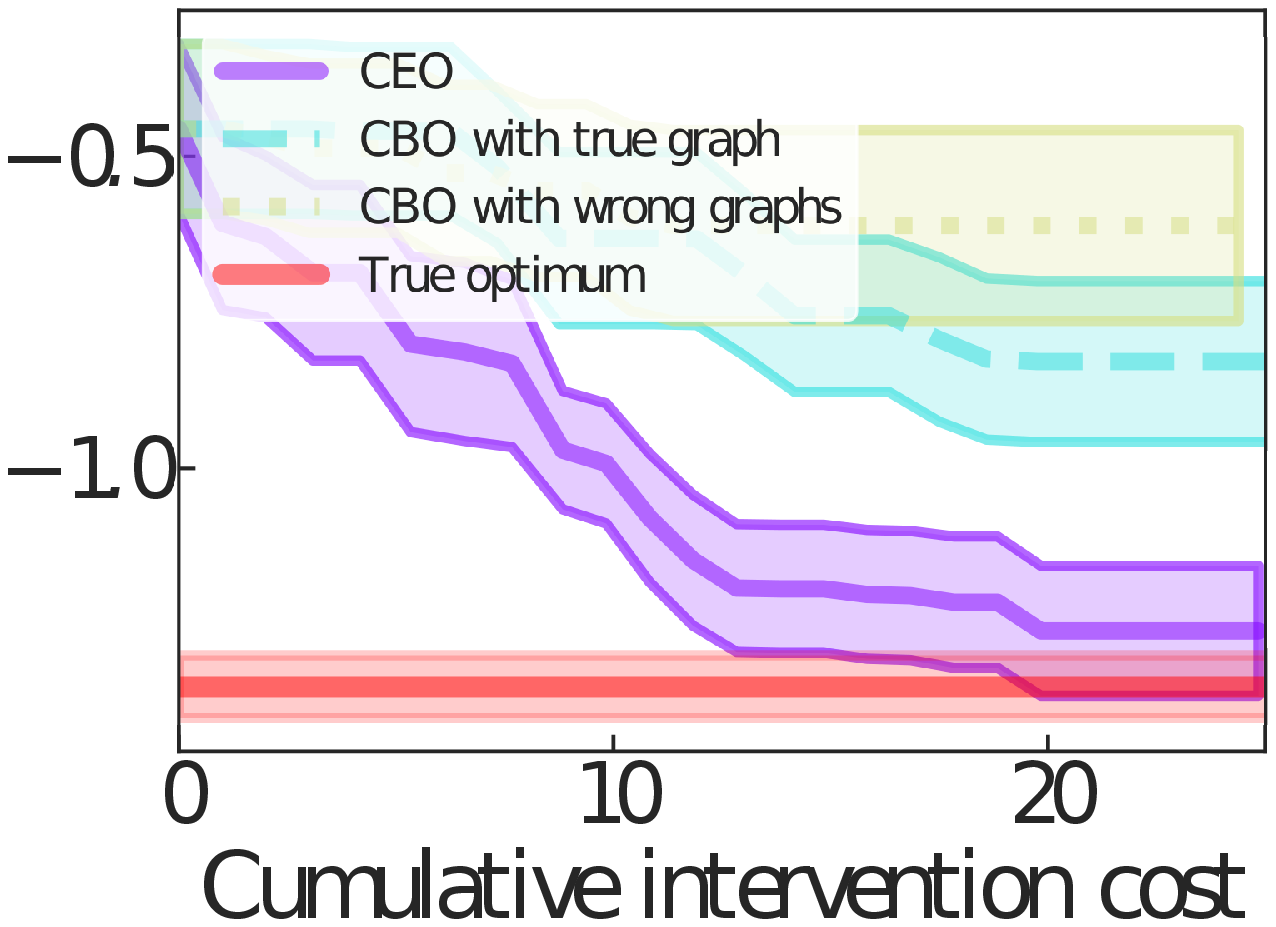}
        \caption{Ext. Epidemiology}
    \end{subfigure}
    \caption{Convergence plot for \ceo and competitor methods across replicates. In the Healthcare and Epidemiology examples, the graph is found immediately by the posterior with the initial $\dataset^{O}$ and $\dataset^{I}$; therefore $\acro{cd-cbo}$ simply reduces to \cbo. Shaded areas are $\pm$ one standard error. We plot $\pm 0.05$ around the global optimum shaded in red. Vertical axis shows best value of the causal effect achieved, horizontal shows \emph{cumulative} intervention cost. \vspace{-10pt} \label{fig:convergence_results}}
\end{figure*}

\begin{wrapfigure}{r}{0.4\textwidth}
    \centering
    \includegraphics{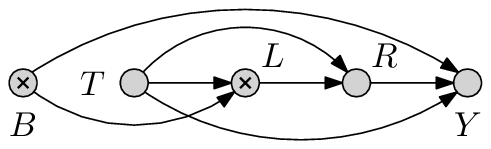}
    \caption{True \DAG used for the epidemiology experiment. For the full support of $P(G)$, and \scm equations, see App. Section J. Crossed variables are non-manipulative. \label{fig:epi_DAG}}
    \vskip -1em
\end{wrapfigure}

\textbf{Epidemiology}
In this example, adapted from \citep{havercroft2012simulating}, the goal is to subministrate doses for two potential treatments, which we denote as $T$ and $R$, (see \citep{havercroft2012simulating} for details) to minimize HIV viral load. The associated \DAG is shown in \cref{fig:epi_DAG}. \cref{fig:convergence_results}(c) shows how, in this more challenging scenario, both competitor methods perform worse. Indeed, the multimodal nature of the causal effects and the high observation noise characterizing this example penalise \cbo and its \ei acquisition function, which is known to be more greedy than entropy based one \citep{frazier2018bayesian}. While this has been observed in a-causal \bo, it is even more problematic when exploring and comparing multiple functions as in \cbo.

\textbf{Extended Epidemiology}: To conclude, we test the algorithm on a larger graph obtained by adding confounders to the Epidemiology \DAG. The resulting \DAG includes 10 nodes, and the associated wrong graphs are created by removing the same edges from the original graph as in the Epidemiology experiment. Even under this situation, as in the previous Epidemiology experiment, we find that \ceo generally performs better than \cbo (with both wrong and true). Indeed, a larger graph implies a larger product in the truncated factorization term; this can imply further overconfidence in the \cbo surrogate model, whereas \ceo averages over multiple plausible hypotheses.

\section{Discussion \& Conclusions}\label{sec:conclusions}

We proposed \ceo, a framework that allows an experimenter to efficiently solve the causal global optimization problem when the graph is unknown. \ceo handles continuous variables with flexible nonparametric relationships, while still allowing for a closed-form posterior over graphs that combines observational and interventional data. Our experiments with synthetic and real-world \DAG{s} show that \ceo allows the experimenter to reach the global optimum with significantly reduced cost compared to \cbo when the graph is unknown, and sometimes even when the graph is known. Our acquisition, \ces, allows to learn the graph only to the extent to which it is useful for optimization, and it is robust to observation noise. 

A limitation of \ceo is its restriction to continuous variables, which inherits from \cbo. Further, a significant computational effort is required to efficiently approximate the \ces objective. However, the formulation of \ces can be seen as an active-learning objective of independent interest, and thus opens promising avenues for future work (e.g. joint causal effect estimation and structure learning). Further, our surrogate model definition could also allow for approximating expectations over the graph (e.g. exploiting the rich literature on MCMC with dynamic programming for structure learning \citep{eaton2007bayesian,zemplenyi2021bayesian}) to help scalability of posterior inference in very large graphs. Future work could attempt to find an optimal or near-optimal experiment design criterion.

\subsubsection*{Acknowledgments}
TD acknowledges support from a UKRI Turing AI acceleration Fellowship [EP/V02678X/1].
\clearpage

\bibliographystyle{abbrvnat}
\bibliography{ceo_icml}

\clearpage
\appendix 
\onecolumn
\section*{Causal Entropy Optimization: Supplementary Material}
\section{Nomenclature}
\label{sec:nomenclature}

\begin{table}[htbp]
\centering
\renewcommand{\arraystretch}{1.2} 
\begin{tabular}{ccl}
\textbf{Symbol} &  & \textbf{Description} \\
 & & \\
$\mat{V}$ & & Observed endogenous variables \\
$\mat{U}$ & & Set of exogenous background variables  \\
$F$ & & Set of deterministic functions \\
$\mat{C}$ & & Non-manipulative variables  \\
$\mat{X}$ & & Manipulative variables  \\
$Y$ & & Output variable  \\
$\mathcal{P}(\mat{X})$ & & Set of all possible interventions which can be performed in the graph \\
$\mathbf{Pa}_{j}^{\mathcal{G}}$ &  & Parents of each variable $V_j \in \V$ given by $\mathcal{G}$\\
$N$ &  &  Number of samples collected from each interventional distribution \\
$\text{pa}_i$ & & Denotes the parents of $V_i$\\
$\XI$ & & One possible intervention set out of a total $|\mathcal{P}(\mat{X})|$ \\
$\nocolexpval$ & & Corresponding values of intervened set of variables $\XI$ \\
$\VNI$ & & Corresponding set of non-intervened variables following intervention $\XI$ \\
$\mathbf{V}_{Y}$ & & Denotes $\VNI \setminus Y$\\
$D(\XI)$ & & Intervention domain  \\
\textbf{ES} & & Exploration set, equivalent to or subset of $\mathcal{P}(\mat{X})$ \\
$\dataI$ & & Samples of manifestations of variables in $\mathcal{G}$ following intervention $\XI$ \\
$\XIstar$ & & Optimal intervention set \\
$\xIstar$ & & Optimal intervention level(s) \\
$\mathcal{G}$ & & True causal graph \\
$\graphvar$ & & Graph latent random variable \\
$P(G)$ & & Discrete prior over causal graphs \\
$R_G$ & & Support of graph distribution \\
$g$ & & One of the possible graphs in $P(G)$ \\
$\mI(\x_{I})$ & & Mean function used in \gptext prior on  $\mathbb{E}[Y \mid \DO{\XI}{\x_{I}}, \graph]$ \\
$\mathbb{I}(\mat{A} ; \mat{B} | \mat{C})$ & & Conditional mutual information between sets $\mat{A}$ and $\mat{B}$ on $\mat{C}$ \\
$\mathbb{H}(p(x))$ & & Entropy if $x$ is discrete, differential entropy if $x$ is continuous \\
$\Co(\XI,\nocolexpval)$ & & Cost of performing intervention $\DO{\XI}{\nocolexpval}$ \\
$\mathcal{N}_x(0,1)\Bigr|_{X=x}$ &  & Evaluate expression (e.g. $\mathcal{N}_x(0,1)$ ) at $X=x$
\end{tabular}
\label{tab:TableOfNotation}
\end{table}

\section{Further discussion on graph posterior and likelihood}\label{sec:graphposterior}

\cref{eq:prior_mean_int} and \cref{eq:prior_var_int} can also be seen as \emph{estimators} of the true average causal effect $\mathbb{E}[Y \mid \DO{\XI}{\x_{I}}, \graph]$ and the variance of the interventional distribution $\mathbb{V}[Y \mid \DO{\XI}{\x_{I}}, \graph]$. Unsurprisingly, the mean estimator minimizes the error for a risk function with MSE loss in this context \citep{horii2021bayesian}. Both estimators are consistent, in the sense that: (1) as $P(G) \rightarrow \delta_{G = \graph}$ and as $\widehat{p}(Y\mid \DO{\XI}{\x_{I}}, \graphvar=g)  \rightarrow p(Y\mid \DO{\XI}{\x_{I}}, \graphvar=g)$ (ensured by do-calculus, when the effect is identifiable).  Further, as $P(G) \rightarrow \delta_{G = \graph}$ (a Dirac mass at $\graph$), the first term in \cref{eq:prior_var_int} converges to $\mathbb{V}_{\hat{p}(Y\mid \DO{\XI}{\x_{I}}, \graph)}[Y] $, while the second term vanishes. 
\paragraph{Posterior convergence}
There are standard conditions in our setting that ensure that the graph is identified. The following assumptions in our setting:
\begin{enumerate}
    \item Causal sufficiency (i.e. no hidden confounders)
    \item All variables (except the target) can be manipulated (i.e. intervened on)
    \item Infinite samples (observational and interventional) can be obtained from each node
    \item Causal minimality; i.e. the joint in line 74 does not Markov-factorize w.r.t. to any sub-graph of $\mathcal{G}$
    \item The support of $P(G)$ includes $\mathcal{G}$
\end{enumerate}
are sufficient to guarantee convergence of the posterior to a Dirac mass on the true graph, in the limit as interventions are performed. This is independent of how interventions are collected. That we can intervene on all nodes also guarantees us identifiability of the interventional Markov-equivalence classes \citep{hauser2015jointly}. Finally, we are not aware of finite sample guarantees that apply to our specific setting.

\paragraph{Graph likelihood full expression}
Denote by $\X^O$ and $\mathbf{y}^O$ the observational inputs and outputs in $\mathcal{D}^{O}$ and by $pa_{j,I}^{g}$ the values for the parents of $V_j$ in $\graphvar = g$ resulting from an intervention on $\XI = \x_{I}^{(i)}$. We can write the likelihood  evaluated at a single interventional point $(\x_{I}^{(i)}, \mathbf{v}_I^{(i)})$ as:
\begin{align}
     & \prod_{\onevarV \in \VNI} \mathcal{N}_{\onevarV}(m_j, \Sigma_j)\Bigr|_{\substack{V_j = v_{j}^{(i)} }} \;\; \text{with} \;\; A_j = k_j(pa_{j,I}^{g}, \X^O)[k_j(\X^O,\X^O) + \sigma_j^2 I]^{-1}  \label{eq:lik_dataI} 
      \\
     & m_j=A_j \mathbf{y}^O , \;\;  \Sigma_j=k_j(pa_{j,I}^{g}, pa_{j,I}^{g}) - A_j k_j(\X^O,pa_{j,I}^{g})
\end{align}
where $k_j$ represent the prior kernel functions of $f_j$ and $m_j$, $\Sigma_j$ correspond to the standard posterior predictive \gptext parameters \citep[p.17]{williams2006gaussian}. Note that, when the intervened variables are parents of $V_j$, the values of $pa_{j,I}^{g}$ are replaced by the $\x_{I}^{(i)}$ values.

\paragraph{Exploring the full space of \DAG{s}}
Methodologies for exploring the full space of \DAG{s} are well-studied (i.e. defining a posterior over a very large space), and return an approximate solution employing sophisticated \mcmc schemes (e.g. over the space of node \emph{orderings}) in the context of Bayesian structure learning \citep{madigan1995bayesian,friedman2003being,kuipers2017partition,pmlr-v124-viinikka20a}. Our setting is complementary to this research: once a sophisticated approximate posterior is provided via representative samples, this can be incorporated in our framework without adjustments. Therefore, on the structure learning side our work is more related to \citep{von2019optimal}, where also a small number of \DAG{s}, continuous $\X$ and flexible nonparametric priors for \sem{s} are considered; however, they are not interested in causal optimization.

\section{Causal Entropy Search acquisition computation}

\subsection{Joint entropy details}
\label{subsec:marginalization_details}

\begin{algorithm}[tb]
\setstretch{1.35}
   \caption{\ces}
   \label{alg:ces}
\begin{algorithmic}[1]
   \STATE {\bfseries Inputs:}  
   \begin{itemize}
      \setlength\itemsep{1pt}
        \item Initial data $\dataset = (\dataset^{O},\dataset^{I})$
       \item Surrogate models: $f_{I}(\nocolexpval)$ for $\XI \in \setint$, which have been fitted on $\dataset$
       \item Current graph posterior: $P(G | \dataset)$
       \item Intervention sets $\setint$
       \item Corresponding acquisition points $\{\nocolexpval^{(s)} \}_{s=1}^{S}$ (using S acquisition points per set).
   \end{itemize}
   \STATE {\bfseries Outputs:} $\XI^{\text{best}}, \x_{I}^{\text{best}}$, maximizers of \cref{eq:ces_acq}
    \FORALL{$\XI \in \setint$}
    \STATE Get samples $\{y_{I}^{\star,(j)} \}_{j=1}^{J}$ approximately distributed from $p(y_{I}^{\star} | \dataset)$ using Thompson sampling: $(\x_{I}^{\star,(j)}, y_{I}^{\star,(j)}) = \argmin(\argmax)_{\nocolexpval} f_{I}^{(j)}(\nocolexpval), \min(\max)_{\nocolexpval} f_{I}^{(j)}(\nocolexpval) $  with $f_{I}^{(j)} \sim  \gp(\mI(\x_{I}), \kI(\x_{I}, \x_{I}')) $  as per \cref{eq_surrogate}. Store each $\x_{I}^{\star,(j)}$,  associated to each $y_{I}^{\star,(j)}$
    \STATE Fit a \kde estimate $\widehat{p}(y_{I}^{\star} | \dataset)$ to the samples $\{y_{I}^{\star,(j)} \}_{j=1}^{J}$
    \STATE Compute $P(\XI = \XI^\star) $ as in \cref{sec:mab}
    \ENDFOR
    \STATE Obtain samples $\{ y_{k}^{\star} \}_{k=1}^{K}$ from  $p(y^\star | \dataset)$ by sampling from the mixture in \cref{eq:pystar}; keeping track of each associated $\x^{\star,k}$
    \STATE Fit \kde estimate $\widehat{p}(y^{\star} | \dataset)$ to the samples $\{y^{\star,k} \}_{k=1}^{K}$
    \STATE \textbf{Note}: up to here, computations do not depend on acquisition points.
    \IF{ $\mathbb{H}[p(G | \dataset)]$ is $ \approx 0$ (i.e. \ceo is sure that it has found the graph)}
    \STATE Approximate $\mathbb{H}(p(y^{\star} | \dataset))$ with the entropy of the \kde estimate $\mathbb{H}(\widehat{p}(y^{\star} | \dataset))$ via quadrature
    \FORALL{$\XI \in \setint$}
    \FORALL{$ \nocolexpval^{(s)} \in \{\nocolexpval^{(s)} \}_{s=1}^{S}$}
    \STATE Update (a copy of) the surrogate model   $f_{I}(\nocolexpval)$ with $\nocolexpval^{(s)} $
    \STATE Generate fantasy observation by performing $\DO{\XI}{\nocolexpval^{(s)}}$: sample $\mathbf{v}_{Y}^{l} \sim p(\mathbf{v}_{Y}^{l} | \text{do}(\nocolexpval^{(s)}))$ using ancestor sampling with \sem functions estimated on $\dataset$   , whereas $y^{l} \sim p(Y | \text{do}(\nocolexpval^{(s)}), \dataset)$ is given by the surrogate model on $Y$ , for $l = 1,\dots,L$.
    \STATE Repeat steps 3 to 9 to get an updated set $ \{ \widehat{p}(y^{\star} | \dataset \cup (\nocolexpval^{(s)}, \mathbf{v}_{Y}^{l}, y^{l})) \}_{l=1}^{L}$
    \STATE Compute the average change in entropy, conditioned on fantasy observations: $\frac{1}{L}\sum_{l=1}^{L} \mathbb{H}(\widehat{p}(y^{\star} | \dataset)) -\mathbb{H}(\widehat{p}(y^{\star} | \dataset \cup (\nocolexpval^{(s)}, \mathbf{v}_{Y}^{l}, y^{l})))  $ 
    \ENDFOR
    \ENDFOR
    \STATE Return $\nocolexpval^{(s)} $ maximizing its change in entropy as computed in step 18
  \ELSIF{$\mathbb{H}[p(G | \dataset)]$ is not $ \approx 0$}
    \STATE Compute joint entropies $ \mathbb{H}[p(y^\star,G  | \dataset )]$ as in \cref{eq:jointeentr} and $y^\star$ samples obtained in step 8
    \STATE Do steps 13 - 20, but for these joint entropies, for all $\nocolexpval^{(s)}, \mathbf{X}_I$:   $\frac{1}{L}\sum_{l=1}^{L} \mathbb{H}[p(y^\star,G  | \dataset )] -   \mathbb{H}[p(y^\star,G  | \dataset \cup (\nocolexpval^{(s)}, \mathbf{v}_{Y}^{l}, y^{l}) )] $
    \ENDIF
    \STATE \textbf{Return}: $\mathbf{x}^{\text{best}}$ as the one associate with larger entropy reduction among the $\{\nocolexpval^{(s)} \}_{s=1}^{S}$, and its associated $\mathbf{X}^{\text{best}}$

\end{algorithmic}
\end{algorithm}

For simplicity we omit conditioning on $\dataset$ on all terms, and write the joint entropy as:
\begin{align}\label{eq:jointeentr}
    \mathbb{H}[p(y^\star,G  )] &= -\sum_G \int \mathrm{d} y^\star p(G,y^\star   )  \log p(G,y^\star  ) \\
 &= -\sum_G \int \mathrm{d} y^\star p(G | y^\star) p(y^\star)  \log p(G |y^\star  )  -  \sum_G \int p(G | y^\star) p(y^\star)  \log p(y^\star) \\
 &= -\sum_G \int \mathrm{d} y^\star p(G | y^\star) p(y^\star)  \log p(G |y^\star  )  -   \int  p(y^\star)  \log p(y^\star) 
\end{align}
We approximate the challenging term (the first) with Monte Carlo samples from our mixture distribution $p(y^\star | \dataset)$ (\cref{eq:pystar}) and keep track of the $\x^\star$ associated with each sample of $y^\star$ to update the graph posterior. More accurate approximations could be considered in future work. The second term we approximate as described in \cref{alg:ces}, which details the complete algorithm for \ces .
\section{Graph posterior evolution plots}\label{sec:posterior_evo}
\begin{figure}[ht!]
    \centering
    \begin{subfigure}[t]{0.4\textwidth}
        \centering
        \includegraphics[width=1\textwidth]{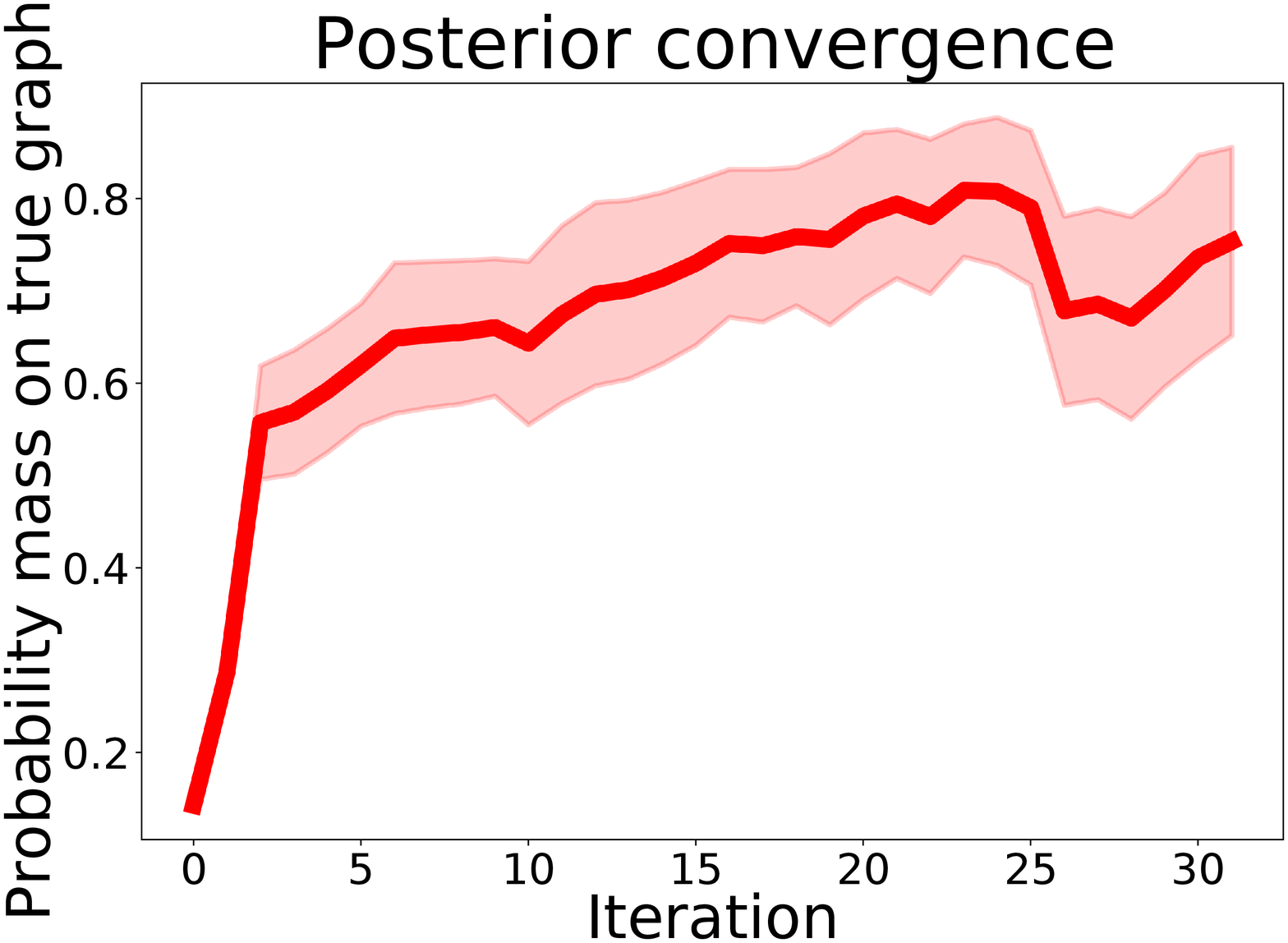}\caption{}
    \end{subfigure}%
    \hfill
    \begin{subfigure}[t]{0.4\textwidth}
        \centering
        \includegraphics[width=1\textwidth]{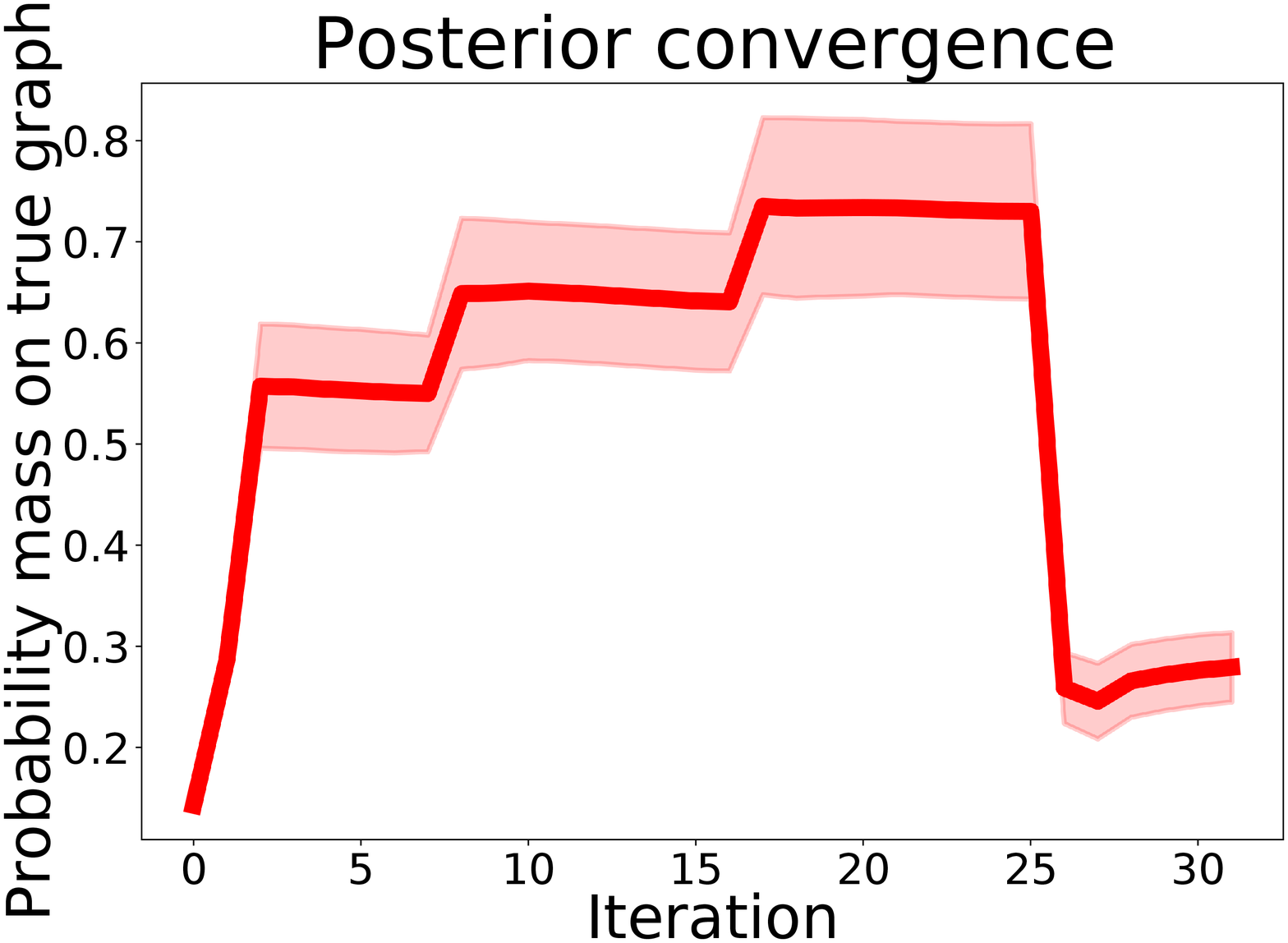}\caption{}
    \end{subfigure} \caption{Evolution of $P(G = \graph)$ across iterations in the experiment using the synthetic graph.  (a) is the posterior as given by \acro{cd-cbo}, whereas (b) is that given by \ceo . As discussed in \cref{sec:experiments}, we found this an interesting example where the true graph is hard to distinguish from an alternative one, but causal optimization can still proceed efficiently. Note that in both healthcare and epidemiology examples we found the graph to be identified easily by our posterior.  \label{fig:convergence_post}}
\end{figure}

 \section{Further discussion on Causal Entropy Search and related literature} Compared to acquisition functions in prior related works \citep{cbo,wang2017max,NIPS2013_f33ba15e}, several important challenges arise specific to our goals as specified by \cref{eq_objective}: to start with, the acquisition used in \cbo cannot be straightforwardly extended to our setting. Firstly, because it is not clear how to incorporate our graph prior in a principled manner. Secondly, as explained in \cref{sec:inf_causal_effects}, we cannot assume that we can observe the causal effect exactly, since observations from the \sem are noisy, and the family of expected-improvement (\ei) acquisitions are known to be inappropriate in this setting \citep{frazier2018bayesian,garnett_bayesoptbook_2022}. 
 We addressed these challenges by introducing an information-theoretic acquisition function. These types of objectives are widely used in \boed  \citep{drovandi2017principles}, active learning \citep{mackay1992information} as well as many \bo approaches \citep{hennig2012entropy,wang2017max,moss2021general}. The information-theoretic approach is often advocated in \bo because, contrasted to approaches like \ei that largely judge optimization performance based \emph{solely} on having found high objective function values, it seeks data that is maximally informative about a variable of interest \citep{garnett_bayesoptbook_2022}. In practice, these approaches can be better suited to challenging noisy problems,  multimodal and non-smooth functions, or optimization with multiple information sources  \citep{frazier2018bayesian}. It is worth keeping in mind that there is no uniformly best acquisition across all settings.

 \textbf{Output space vs input space ES}
  We decided to frame causal optimization as learning about $y^\star$ rather than $\x^\star$. The question of whether to perform output-space ES (what we do) versus input-space ES has been studied before in \bo; see \citep[\S 6]{garnett_bayesoptbook_2022} and \citep{moss2021general}. However, a crucial difficulty is added in causal optimization: there are \emph{multiple} surrogate models, each with different input space and dimensionality; one for each intervention set $\mathbf{X}_I \in \setint$ we want to consider. Having multiple surrogates is similar to the setting of multi-task \bo  \citep{NIPS2013_f33ba15e}; however (1) our tasks do not share input space and (2) we do not know which task actually contains the global optimum. Therefore, while in principle learning about $\mathbf{x}^\star$ could be done, it would be computationally expensive and inference-wise challenging to define a distribution over a variable with varying (and potentially large) dimensionality. On the other hand, all causal effects share $Y$, which is one dimensional. Future work could consider inference techniques for dimension-varying parameters like reversible-jump \mcmc \citep{green1995reversible}.

\section{Further discussion on surrogate models}
\label{sec:appendix_surrogate_discussion}

\paragraph{A surrogate model for each graph?} Notice that one can think of a very different way to incorporate uncertainty: we could define distinct surrogate models for each $g$ in the support $R_G$ of $P(G)$. We do not take this approach for several reasons: (1) it would not extend to a setting where we need to approximate expectations under $P(G)$ with sampling (hence it is not scalable in this sense); (2) it would require $|R_G| \cdot |\setint|$ kernel hyperparameters to store and update at every iteration; (3) finally, modelling the true causal effect by marginalizing the graph in this approach would lead to a weighted mixture of \gptext{s}, rendering inference and \bo complicated.

\section{Further derivations}

The variance term in \cref{eq:prior_var_int}  is computed as
\begin{equation}
\mathbb{V}[\hatexpectation{}{Y\mid \DO{\XI}{\x_{I}}, \graphvar}]
= \expectation{}{\left [ \widehat{\mathbb{E}}[Y | \DO{\XI}{\x_I}, G] \right ]^2} - (\mI(\x_{I}))^2 .
\end{equation} 


\section{Belief over the optimal set}
\label{sec:mab}
We compute $P(\XI = \XI^\star)$ as:
\begin{equation}
    P(\XI = \XI^\star) = \frac{e^{\mu_{I}^\star + \beta \cdot  \sigma_{I}^\star }}{\sum_{I^\prime}e^{\mu_{I^\prime}^\star + \beta \cdot  \sigma_{I^\prime}^\star }}
\end{equation}
when maximizing a causal effect, with signs reversed when minimizing. Here, $\mu_{I}^\star$ is the minimum of the mean function of surrogate model $f_I$, and $\sigma_{I}^\star$ the standard deviation of the univariate Gaussian with mean at that point. We did not explore adaptive tuning of $\beta$ which we fixed to $0.1$. Guarantees for UCB can be found in e.g. \citep{garnett_bayesoptbook_2022}.

\subsection{Parametric assumptions}
     Firstly, the additive Gaussian noise assumption (let us call it AGN- additive Gaussian noise) combined with GPs on the structural equations allows us to define the graph likelihood in closed form. This also implies that GP posteriors on the structural functions are available in closed form. When AGN does not hold, one would need to resort to the literature on Variational GPs, which allows one to perform inference with GPs with non-Gaussian likelihoods. Secondly, the AGN assumption also allows us to deal with noisy Bayesian optimization \citep{garnett_bayesoptbook_2022}. Recall that, even if we knew the graph, we do not get to observe  $\mathbb{E}[Y \mid \text{do} ({\mathbf{X}_{I}} = {\mathbf{x}_{I}}), G]$, but can only get samples from $p(Y \mid \text{do} ({\mathbf{X}_{I}} = {\mathbf{x}_{I}}), G)$. BO with non-Gaussian noise is an active area of research, see e.g. [16] who allow for sub-Gaussian likelihood noises. The optimization becomes more sophisticated, and note that CEO (and CBO) needs to deal with multiple surrogate models (one for each $\mathbf{X}_{I}$). Therefore, extending these methods from the BO setting to CBO and CEO is a future area of research. \\ Finally, beyond the previously discussed additive Gaussian noise assumption, in our experiments we used radial basis function kernels, consistently with previous works on CBO. These were appropriate kernels for the SCMs we studied in the experiments; however, for nonstationary causal effect functions one could use nonstationary GP kernels.

\section{Synthetic: graphs and true \sem}
\label{sec:synthetic_graph_priors}
For the \sem of the synthetic graph, see the Supplementary material of \citep{cbo}. For all experiments, we used standard radial basis function (RBF) kernels, and optimized hyperparameters with type $II$ maximum likelihood.
\begin{figure}[ht!]
    \centering
    \begin{subfigure}[t]{0.33\textwidth}
        \centering
        \includegraphics[width=0.7\textwidth]{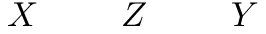}
        \caption{}
    \end{subfigure}%
    \hfill
    \begin{subfigure}[t]{0.33\textwidth}
        \centering
        \includegraphics[width=0.7\textwidth]{figures/dag_priors/dag_prior_2_eps.eps}
        \caption{}
    \end{subfigure}%
    \hfill
    \begin{subfigure}[t]{0.33\textwidth}
        \centering
        \includegraphics[width=0.7\textwidth]{figures/dag_priors/dag_prior_3_eps.eps}
        \caption{}
    \end{subfigure}%
    \\ 
    \vskip 20pt
    \begin{subfigure}[t]{0.33\textwidth}
        \centering
        \includegraphics[width=0.7\textwidth]{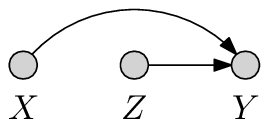}
        \caption{}
    \end{subfigure}%
    \hfill
    \begin{subfigure}[t]{0.33\textwidth}
        \centering
        \includegraphics[width=0.7\textwidth]{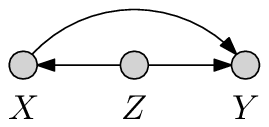}
        \caption{}
    \end{subfigure}%
    \hfill
    \begin{subfigure}[t]{0.33\textwidth}
        \centering
        \includegraphics[width=0.7\textwidth]{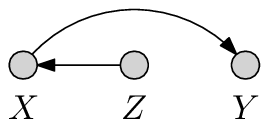}
        \caption{}
    \end{subfigure}%
    \\ 
    \vskip 20pt
    \begin{subfigure}[t]{0.33\textwidth}
        \centering
        \fcolorbox{green}{white}{\includegraphics[width=0.7\textwidth]{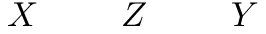}}
        \caption{\label{fig:synthetic_true_dag}}
    \end{subfigure}%
    \caption{Graph prior $P(G)$ used for the synthetic experiment s.t. $|R_G| = 6$. The true \DAG is shown in \cref{fig:synthetic_true_dag}.
    }
\end{figure}
\FloatBarrier

\section{Epidemiology: graphs and true \sem}
\label{sec:epi_graph_priors}
We use a modified (more challenging and nonliner) version of the \sem only partially specified in \citep{havercroft2012simulating}:
\begin{align}
    B &= \mathcal{U}[-1,1] \\
    T &= \mathcal{U}[4,8] \\
    L &= \text{expit}(0.5 \cdot T + U) \\
    R &= 4 + L \cdot T \\
    Y &= 0.5 + \cos (4 \cdot T) + \sin (- L + 2 \cdot R) + U + \epsilon \qquad \text{with} ~~\epsilon ~~\sim \mathcal{N}(0,1)
\end{align}
\begin{figure}[ht!]
    \centering
    \begin{subfigure}[t]{0.4\textwidth}
        \centering
        \includegraphics[width=\textwidth]{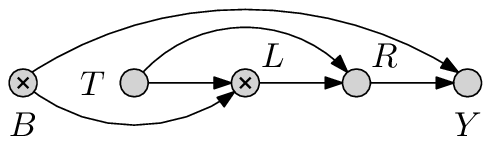}
        \caption{}
    \end{subfigure}%
    \quad
    \begin{subfigure}[t]{0.4\textwidth}
        \centering
        \includegraphics[width=\textwidth]{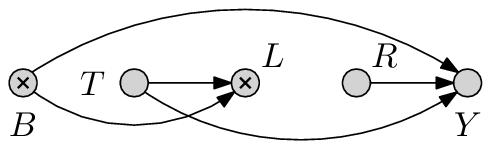}
        \caption{}
    \end{subfigure}%
    \\
    \vskip 20pt
    \begin{subfigure}[t]{0.4\textwidth}
        \centering
        \includegraphics[width=\textwidth]{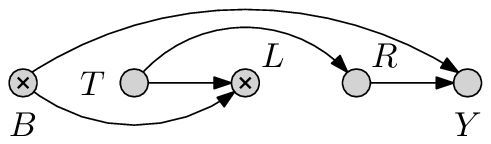}
        \caption{}
    \end{subfigure}%
    \quad
    \begin{subfigure}[t]{0.4\textwidth}
        \centering
        \fcolorbox{green}{white}{\includegraphics[width=\textwidth]{figures/epi_priors/epi_example.eps}}
        \caption{\label{fig:epi_true_dag}}
    \end{subfigure}%
    \caption{Graph prior $P(G)$ used for the epidemiology experiment s.t. $|R_G| = 3$. The true \DAG is shown in \cref{fig:epi_true_dag}.
    }
\end{figure}
\FloatBarrier

\section{Healthcare: : graphs and true \sem}
\label{sec:health_graph_priors}
For the \sem see the Supplementary material of \citep{cbo}. In this example, we had initial points $|\dataset^{{I}}|= 2$.

\begin{figure}[ht!]
    \centering
    \begin{subfigure}[t]{0.4\textwidth}
        \centering
        \includegraphics[width=\textwidth]{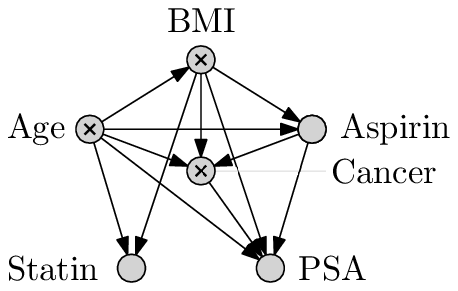}
        \caption{}
    \end{subfigure}%
    \qquad
    \begin{subfigure}[t]{0.4\textwidth}
        \centering
        \includegraphics[width=\textwidth]{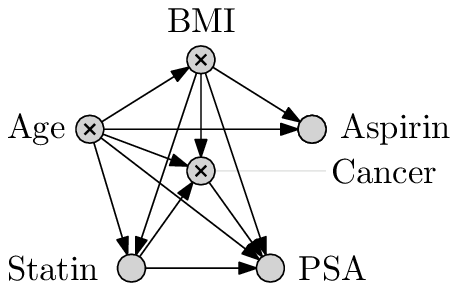}
        \caption{}
    \end{subfigure}%
    \\
    \vskip 20pt
    \begin{subfigure}[t]{0.4\textwidth}
        \centering
        \includegraphics[width=\textwidth]{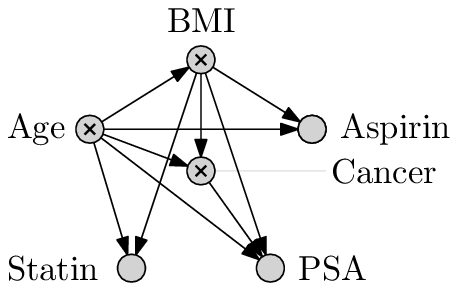}
        \caption{}
    \end{subfigure}%
    \qquad
    \begin{subfigure}[t]{0.4\textwidth}
        \centering
        \includegraphics[width=\textwidth]{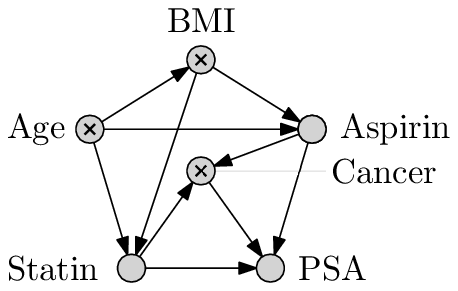}
        \caption{}
    \end{subfigure}%
    \\
    \vskip 20pt
    \begin{subfigure}[t]{0.4\textwidth}
        \centering
        \fcolorbox{green}{white}{\includegraphics[width=\textwidth]{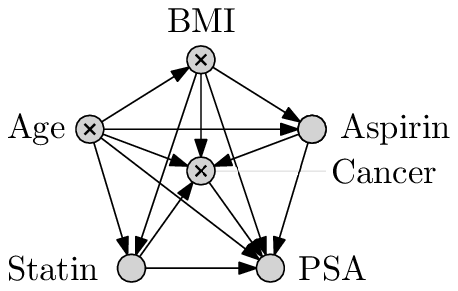}}
        \caption{\label{fig:health_true_dag}}
    \end{subfigure}%
    \caption{Graph prior $P(G)$ used for the health experiment s.t. $|R_G| = 4$. The true \DAG is shown in \cref{fig:health_true_dag}.
    }
\end{figure}
\FloatBarrier

\section{Entropy in Causality}
There is a large literature on the use of information theoretic concepts with applications to causal inference, discovery and counterfactual reasoning. The information-geometric approach to causal discovery \citep{janzing2012information} attempts to lift the strict conditional independence requirement of classical algorithms like PC \citep{spirtes2000causation}, by defining independence as orthogonality in information space. They find that in important examples this induces a desirable asymmetry between cause and effect. \citet{miklin2017entropic} find use in entropies between variables of a causal Bayesian network in deriving so-called causal inequalities, which are bounds on certain quantities that characterize the behaviour of the underlying system represented by the CBN. Recently, \citet{kocaoglu2017entropic} introduced a framework, ``Entropic Causal Inference'', where the causal direction between two categorical variable can be discerned from observational data based on an interesting condition on the entropy of the exogenous variable.   

\section{Computational Complexity}
We provide here more details on computational complexity. \\
 We have now added a paragraph in Section 3.3 about complexity and a new Section in the Supplement with more details The complexity of CEO is driven by the computation of CES. The parameters influencing these are: $|\mathbf{ES}|$, $\max_{I} |\mathbf{X}_{I}|$, the number of acquisition points $N$ (i.e., how many values of the intervened variable to consider, assuming the same number for all $\mathbf{X}_{I}$). Therefore, the total complexity of the acquisition is $\mathcal{O}(N \cdot \text{CES}(\mathbf{x}) \cdot \sum_{\mathbf{X}_{I} \in \mathbf{ES}} |\mathbf{X}_{I}|  )$. Here, $\text{CES}(\mathbf{x})$ denotes the time needed to compute CES for m specific value $\mathbf{x}$, regardless of its corresponding $\mathbf{X}_{I}$. Note this is valid for CBO also, replacing $\text{CES}(\mathbf{x})$ with $\text{CEI}(\mathbf{x})$, the acquisition used by CBO. However,$\text{CEI}(\mathbf{x})$ is cheaper than $\text{CES}(\mathbf{x})$. The biggest reason in practice is that CEI operations can be more easily vectorized. CES operations however can be parallelized over both $\mathbf{ES}$ and values of $\mathbf{x}_{I}$. Our implementation indeed uses KDEs, but it does so only to estimate univariate marginal distributions, which is not very expensive (no curse of dimensionality). \textbf{Graph sizes}: Since in our setting the number of nodes is not too large, there is a tractable number of graphs that can be enumerated, therefore omitted in the above Big-O notation. As mentioned in \cref{sec:graphposterior}, larger spaces of graphs could be explored in future work by sampling with MCMC as common in the structure learning literature, losing however our closed form expectations over the graph. Work that explores the full space of DAGs with many nodes will necessarily introduce additional approximations. We also discussed scalability in the limitations in Section 6. It is a general limitation, not restricted to CEO, that in causal global optimization problems one needs to train $|\mathbf{ES}|$ GPs ( in general $|\mathbf{ES}|$ will scale as $2^{|\mathbf{X}|}$), which does not scale with large graphs (here exact GP inference is cubic in the number of collected interventions).

\section{Further experimental details and additional results}\label{sec:additionalresults}
\begin{figure}
    \centering
    \includegraphics[width=0.6\textwidth]{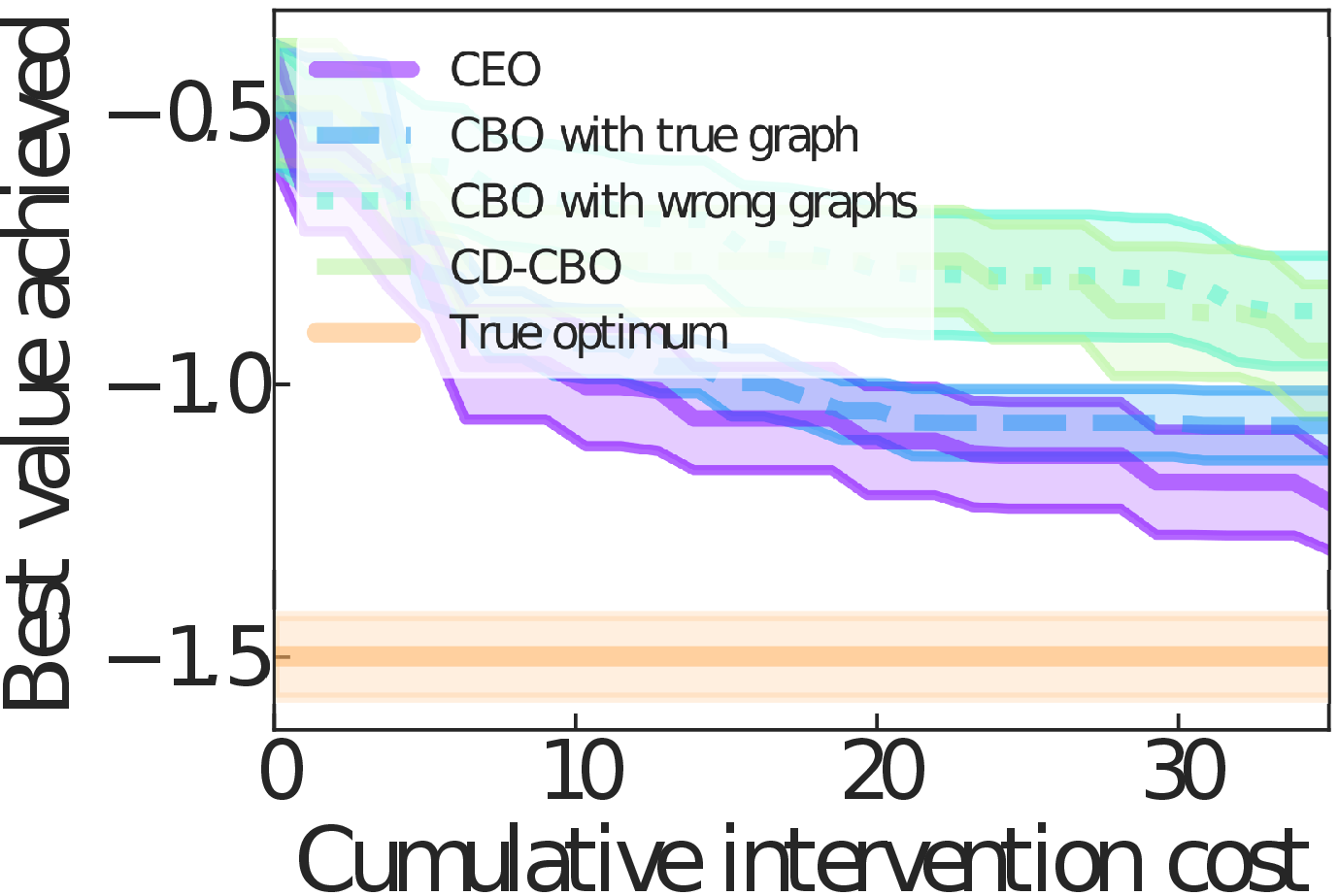}
    \caption{Additional results for a comparision with CD-CBO on the Epidemiology example.}
    \label{fig:add_results}
\end{figure}

We provide here additional experimental results. In \cref{tab:table_gaps_additional} and \cref{fig:add_results} we show that, as in our previous experiment, CD-CBO performs worse than \our, and in this case slightly better than CBO on the true graph. Since as we mentioned in \cref{sec:experiments}, in this example we initially found that our acquisition finds the graph too fast (i.e., at initialization), and therefore it would not be possible to compare to CD-CBO, to provide this additional comparison we updated all graph posteriors (of all methods) only with interventions and not with observational data. This only amplifies the signal between the difference among CD-CBO and CBO, and has no other side-effects on the performance comparison; note that if the graph is found immediately, then CD-CBO is CBO. 

\begin{table}[t!]
\centering
\caption{Average \acro{gap} $\pm$ one standard error computed across 12 replicates initialized with different $\dataI$. Higher values are better, and $0 \leq$ \acro{gap} $\leq 1$. The best result for each experiment in bold. \\}
\label{tab:table_gaps_additional}
\begin{tabular}{lcc}
\toprule
Method &  Synthetic & Epidemiology  (\cref{fig:epi_DAG})  \\
\midrule
$\our$ & \textbf{0.75} $\pm 0.03$ & $0.58 \pm 0.05$ \\
$\cbo$ w. $G = \graph$ & $0.66 \pm 0.04$ & \textbf{0.62}$ \pm 0.02$ \\
$\cbo$ w. $G \neq \mathcal{G}$ & $0.59 \pm 0.035$ & $0.46 \pm 0.05$ \\
$\cdcbo$ & $0.39 \pm 0.04$ & $0.52 \pm 0.04$\\
\bottomrule
\end{tabular}
\end{table}

\end{document}

%% file: macros.tex
\newcommand{\eg}{e.g.\xspace}

\newcommand{\iid}{i.i.d.\xspace}

\newcommand{\acro}[1]{\textsc{#1}\xspace}

\newcommand{\gptext}{\acro{gp}}

\newcommand{\mcmc}{\acro{mcmc}}

\DeclareMathOperator*{\argmax}{arg\,max}
\DeclareMathOperator*{\argmin}{arg\,min}

\newcommand\x{\mathbf{x}}

\definecolor{dorange}{RGB}{223,116,0}

\definecolor{dgreen}{RGB}{5,118,0}

\newcommand{\bo}{\acro{bo}}

\newcommand{\ceo}{\acro{ceo}}
\newcommand{\ces}{\acro{ces}}
\newcommand{\boed}{\acro{boed}}
\newcommand{\cbo}{\acro{cbo}}
\newcommand{\cdcbo}{\acro{cd}-\acro{cbo}}

\newcommand{\cbn}{\acro{cbn}}
\newcommand{\our}{\acro{ceo}}

\newcommand{\DAG}{\acro{dag}}
\newcommand{\kde}{\acro{kde}}

\newcommand{\ei}{\acro{ei}}
\newcommand{\sem}{\acro{sem}}
\newcommand{\mi}{\acro{mi}}
\newcommand{\scm}{\acro{scm}}

\newcommand{\psa}{\acro{psa}}

\newcommand{\bmi}{\acro{bmi}}

\newcommand{\mec}{\acro{mec}}
\newcommand{\cd}{\acro{cd}}

\newcommand{\Co}{\textsc{Co}}

%% file: notation.tex
\newcommand{\mat}[1]{\mathbf{#1}}
\renewcommand{\vec}[1]{ \mathbf{#1} } 

\newcommand{\gp}{\mathcal{GP}}

\newcommand{\dataset}{\mathcal{D}}
\newcommand{\X}{\mat{X}}
\newcommand{\V}{\mat{V}}
\newcommand{\C}{\mat{C}}
\newcommand{\U}{\mat{U}}

\newcommand{\expectation}[2]{ \mathbb{E}_{#1}{\left[#2\right]} }
\newcommand{\hatexpectation}[2]{ \hat{\mathbb{E}}_{#1}{\left[#2\right]} }

\newcommand{\DO}[2]{\text{do} (#1 = #2 )}

\newcommand{\graph}{\mathcal{G}}

\newcommand{\graphvar}{G}
\newcommand{\suppG}{R_{\graphvar}}

\newcommand{\indexI}{I}
\newcommand{\XI}{\X_{\indexI}}
\newcommand{\onevarV}{V_{j}}

\newcommand{\VNI}{\V_{\indexI}}
\newcommand{\valVNI}{\vec{v}_{\indexI}}
\newcommand{\setint}{\acro{\textbf{es}}}

\newcommand{\valVNINY}{\vec{v}_{\indexI_Y}}
\newcommand{\dataI}{\dataset}

\newcommand{\fI}{f_{I}} 
\newcommand{\mI}{m_{I}}
\newcommand{\kI}{k_{I}}

\newcommand{\fj}{f_j}
\newcommand{\XIstar}{\XI^{\star}}
\newcommand{\xIstar}{\x_I^{\star}}

\newcommand{\bestacqceo}[1]{\alpha_{\textsc{CEO}}^{I\star}} %
\newcommand{\nocolexpval}{\x_{I}}
